\documentclass{article}

\usepackage{color}
\usepackage[usernames,dvipsnames,table]{xcolor}
\definecolor{shadecolor}{gray}{0.9}

\definecolor{iospeccolor}{RGB}{243,191,187}
\definecolor{intentcolor}{RGB}{159,203,174}

\usepackage[round]{natbib}

\usepackage[accepted]{icml2024}

\usepackage{amsmath}
\usepackage[utf8]{inputenc} %
\usepackage[T1]{fontenc}    %
\usepackage[framemethod=tikz]{mdframed}

\usepackage{url}            %
\usepackage{booktabs}       %
\usepackage{amsfonts}       %
\usepackage{nicefrac}       %
\usepackage{microtype}      %

\usepackage{amsthm} 
\usepackage{graphicx}
\usepackage{subcaption}
\usepackage{tikz}
\usetikzlibrary{shapes,arrows}
\usepackage[colorlinks,linktoc=all]{hyperref}

\usepackage{caption}
\usepackage{multirow}
\usepackage{wrapfig}
\usepackage{amsfonts, amsmath, amssymb}
\usepackage[nameinlink,capitalize]{cleveref}
\usepackage{titlesec}
\usepackage{enumitem}
\usepackage{subcaption}
\usepackage{soul}
\usepackage{bm}
\usepackage{wrapfig}
\usepackage{listings}
\usepackage[title]{appendix}
\usepackage[bottom]{footmisc}
\usepackage[makeroom]{cancel}
\usepackage{csquotes}
\usepackage{svg}
\usepackage{arydshln}
\usepackage{xspace}
\usepackage[draft]{minted}
\definecolor{mydarkblue}{rgb}{0,0.1,0.45}
\definecolor{LightGray}{gray}{0.9}
\definecolor{deepblue}{rgb}{0,0,0.5}
\definecolor{deepred}{rgb}{0.6,0,0}
\definecolor{deepgreen}{rgb}{0,0.5,0}
\definecolor{darkgreen}{RGB}{43,163,39}
\definecolor{bluesquare}{rgb}{126,166,224}

\hypersetup{
   colorlinks=true,
   linkcolor=mydarkblue,
   citecolor=mydarkblue,
   filecolor=mydarkblue,
   urlcolor=mydarkblue}

\makeatletter
\def\PYGborland@reset{\let\PYGborland@it=\relax \let\PYGborland@bf=\relax%
    \let\PYGborland@ul=\relax \let\PYGborland@tc=\relax%
    \let\PYGborland@bc=\relax \let\PYGborland@ff=\relax}
\def\PYGborland@tok#1{\csname PYGborland@tok@#1\endcsname}
\def\PYGborland@toks#1+{\ifx\relax#1\empty\else%
    \PYGborland@tok{#1}\expandafter\PYGborland@toks\fi}
\def\PYGborland@do#1{\PYGborland@bc{\PYGborland@tc{\PYGborland@ul{%
    \PYGborland@it{\PYGborland@bf{\PYGborland@ff{#1}}}}}}}
\def\PYGborland#1#2{\PYGborland@reset\PYGborland@toks#1+\relax+\PYGborland@do{#2}}

\@namedef{PYGborland@tok@w}{\def\PYGborland@tc##1{\textcolor[rgb]{0.73,0.73,0.73}{##1}}}
\@namedef{PYGborland@tok@c}{\let\PYGborland@it=\textit\def\PYGborland@tc##1{\textcolor[rgb]{0.00,0.53,0.00}{##1}}}
\@namedef{PYGborland@tok@cp}{\def\PYGborland@tc##1{\textcolor[rgb]{0.00,0.50,0.50}{##1}}}
\@namedef{PYGborland@tok@cs}{\let\PYGborland@bf=\textbf\def\PYGborland@tc##1{\textcolor[rgb]{0.00,0.53,0.00}{##1}}}
\@namedef{PYGborland@tok@s}{\def\PYGborland@tc##1{\textcolor[rgb]{0.00,0.00,1.00}{##1}}}
\@namedef{PYGborland@tok@sc}{\def\PYGborland@tc##1{\textcolor[rgb]{0.50,0.00,0.50}{##1}}}
\@namedef{PYGborland@tok@m}{\def\PYGborland@tc##1{\textcolor[rgb]{0.00,0.00,1.00}{##1}}}
\@namedef{PYGborland@tok@k}{\let\PYGborland@bf=\textbf\def\PYGborland@tc##1{\textcolor[rgb]{0.00,0.00,0.50}{##1}}}
\@namedef{PYGborland@tok@ow}{\let\PYGborland@bf=\textbf}
\@namedef{PYGborland@tok@nt}{\let\PYGborland@bf=\textbf\def\PYGborland@tc##1{\textcolor[rgb]{0.00,0.00,0.50}{##1}}}
\@namedef{PYGborland@tok@na}{\def\PYGborland@tc##1{\textcolor[rgb]{1.00,0.00,0.00}{##1}}}
\@namedef{PYGborland@tok@gh}{\def\PYGborland@tc##1{\textcolor[rgb]{0.60,0.60,0.60}{##1}}}
\@namedef{PYGborland@tok@gu}{\def\PYGborland@tc##1{\textcolor[rgb]{0.67,0.67,0.67}{##1}}}
\@namedef{PYGborland@tok@gd}{\def\PYGborland@tc##1{\textcolor[rgb]{0.00,0.00,0.00}{##1}}\def\PYGborland@bc##1{{\setlength{\fboxsep}{0pt}\colorbox[rgb]{1.00,0.87,0.87}{\strut ##1}}}}
\@namedef{PYGborland@tok@gi}{\def\PYGborland@tc##1{\textcolor[rgb]{0.00,0.00,0.00}{##1}}\def\PYGborland@bc##1{{\setlength{\fboxsep}{0pt}\colorbox[rgb]{0.87,1.00,0.87}{\strut ##1}}}}
\@namedef{PYGborland@tok@gr}{\def\PYGborland@tc##1{\textcolor[rgb]{0.67,0.00,0.00}{##1}}}
\@namedef{PYGborland@tok@ge}{\let\PYGborland@it=\textit}
\@namedef{PYGborland@tok@gs}{\let\PYGborland@bf=\textbf}
\@namedef{PYGborland@tok@gp}{\def\PYGborland@tc##1{\textcolor[rgb]{0.33,0.33,0.33}{##1}}}
\@namedef{PYGborland@tok@go}{\def\PYGborland@tc##1{\textcolor[rgb]{0.53,0.53,0.53}{##1}}}
\@namedef{PYGborland@tok@gt}{\def\PYGborland@tc##1{\textcolor[rgb]{0.67,0.00,0.00}{##1}}}
\@namedef{PYGborland@tok@err}{\def\PYGborland@tc##1{\textcolor[rgb]{0.65,0.09,0.09}{##1}}\def\PYGborland@bc##1{{\setlength{\fboxsep}{0pt}\colorbox[rgb]{0.89,0.82,0.82}{\strut ##1}}}}
\@namedef{PYGborland@tok@kc}{\let\PYGborland@bf=\textbf\def\PYGborland@tc##1{\textcolor[rgb]{0.00,0.00,0.50}{##1}}}
\@namedef{PYGborland@tok@kd}{\let\PYGborland@bf=\textbf\def\PYGborland@tc##1{\textcolor[rgb]{0.00,0.00,0.50}{##1}}}
\@namedef{PYGborland@tok@kn}{\let\PYGborland@bf=\textbf\def\PYGborland@tc##1{\textcolor[rgb]{0.00,0.00,0.50}{##1}}}
\@namedef{PYGborland@tok@kp}{\let\PYGborland@bf=\textbf\def\PYGborland@tc##1{\textcolor[rgb]{0.00,0.00,0.50}{##1}}}
\@namedef{PYGborland@tok@kr}{\let\PYGborland@bf=\textbf\def\PYGborland@tc##1{\textcolor[rgb]{0.00,0.00,0.50}{##1}}}
\@namedef{PYGborland@tok@kt}{\let\PYGborland@bf=\textbf\def\PYGborland@tc##1{\textcolor[rgb]{0.00,0.00,0.50}{##1}}}
\@namedef{PYGborland@tok@sa}{\def\PYGborland@tc##1{\textcolor[rgb]{0.00,0.00,1.00}{##1}}}
\@namedef{PYGborland@tok@sb}{\def\PYGborland@tc##1{\textcolor[rgb]{0.00,0.00,1.00}{##1}}}
\@namedef{PYGborland@tok@dl}{\def\PYGborland@tc##1{\textcolor[rgb]{0.00,0.00,1.00}{##1}}}
\@namedef{PYGborland@tok@sd}{\def\PYGborland@tc##1{\textcolor[rgb]{0.00,0.00,1.00}{##1}}}
\@namedef{PYGborland@tok@s2}{\def\PYGborland@tc##1{\textcolor[rgb]{0.00,0.00,1.00}{##1}}}
\@namedef{PYGborland@tok@se}{\def\PYGborland@tc##1{\textcolor[rgb]{0.00,0.00,1.00}{##1}}}
\@namedef{PYGborland@tok@sh}{\def\PYGborland@tc##1{\textcolor[rgb]{0.00,0.00,1.00}{##1}}}
\@namedef{PYGborland@tok@si}{\def\PYGborland@tc##1{\textcolor[rgb]{0.00,0.00,1.00}{##1}}}
\@namedef{PYGborland@tok@sx}{\def\PYGborland@tc##1{\textcolor[rgb]{0.00,0.00,1.00}{##1}}}
\@namedef{PYGborland@tok@sr}{\def\PYGborland@tc##1{\textcolor[rgb]{0.00,0.00,1.00}{##1}}}
\@namedef{PYGborland@tok@s1}{\def\PYGborland@tc##1{\textcolor[rgb]{0.00,0.00,1.00}{##1}}}
\@namedef{PYGborland@tok@ss}{\def\PYGborland@tc##1{\textcolor[rgb]{0.00,0.00,1.00}{##1}}}
\@namedef{PYGborland@tok@mb}{\def\PYGborland@tc##1{\textcolor[rgb]{0.00,0.00,1.00}{##1}}}
\@namedef{PYGborland@tok@mf}{\def\PYGborland@tc##1{\textcolor[rgb]{0.00,0.00,1.00}{##1}}}
\@namedef{PYGborland@tok@mh}{\def\PYGborland@tc##1{\textcolor[rgb]{0.00,0.00,1.00}{##1}}}
\@namedef{PYGborland@tok@mi}{\def\PYGborland@tc##1{\textcolor[rgb]{0.00,0.00,1.00}{##1}}}
\@namedef{PYGborland@tok@il}{\def\PYGborland@tc##1{\textcolor[rgb]{0.00,0.00,1.00}{##1}}}
\@namedef{PYGborland@tok@mo}{\def\PYGborland@tc##1{\textcolor[rgb]{0.00,0.00,1.00}{##1}}}
\@namedef{PYGborland@tok@ch}{\let\PYGborland@it=\textit\def\PYGborland@tc##1{\textcolor[rgb]{0.00,0.53,0.00}{##1}}}
\@namedef{PYGborland@tok@cm}{\let\PYGborland@it=\textit\def\PYGborland@tc##1{\textcolor[rgb]{0.00,0.53,0.00}{##1}}}
\@namedef{PYGborland@tok@cpf}{\let\PYGborland@it=\textit\def\PYGborland@tc##1{\textcolor[rgb]{0.00,0.53,0.00}{##1}}}
\@namedef{PYGborland@tok@c1}{\let\PYGborland@it=\textit\def\PYGborland@tc##1{\textcolor[rgb]{0.00,0.53,0.00}{##1}}}

\makeatother

\makeatletter
\def\PYG@reset{\let\PYG@it=\relax \let\PYG@bf=\relax%
    \let\PYG@ul=\relax \let\PYG@tc=\relax%
    \let\PYG@bc=\relax \let\PYG@ff=\relax}
\def\PYG@tok#1{\csname PYG@tok@#1\endcsname}
\def\PYG@toks#1+{\ifx\relax#1\empty\else%
    \PYG@tok{#1}\expandafter\PYG@toks\fi}
\def\PYG@do#1{\PYG@bc{\PYG@tc{\PYG@ul{%
    \PYG@it{\PYG@bf{\PYG@ff{#1}}}}}}}
\def\PYG#1#2{\PYG@reset\PYG@toks#1+\relax+\PYG@do{#2}}

\@namedef{PYG@tok@w}{\def\PYG@tc##1{\textcolor[rgb]{0.73,0.73,0.73}{##1}}}
\@namedef{PYG@tok@c}{\let\PYG@it=\textit\def\PYG@tc##1{\textcolor[rgb]{0.24,0.48,0.48}{##1}}}
\@namedef{PYG@tok@cp}{\def\PYG@tc##1{\textcolor[rgb]{0.61,0.40,0.00}{##1}}}
\@namedef{PYG@tok@k}{\let\PYG@bf=\textbf\def\PYG@tc##1{\textcolor[rgb]{0.00,0.50,0.00}{##1}}}
\@namedef{PYG@tok@kp}{\def\PYG@tc##1{\textcolor[rgb]{0.00,0.50,0.00}{##1}}}
\@namedef{PYG@tok@kt}{\def\PYG@tc##1{\textcolor[rgb]{0.69,0.00,0.25}{##1}}}
\@namedef{PYG@tok@o}{\def\PYG@tc##1{\textcolor[rgb]{0.40,0.40,0.40}{##1}}}
\@namedef{PYG@tok@ow}{\let\PYG@bf=\textbf\def\PYG@tc##1{\textcolor[rgb]{0.67,0.13,1.00}{##1}}}
\@namedef{PYG@tok@nb}{\def\PYG@tc##1{\textcolor[rgb]{0.00,0.50,0.00}{##1}}}
\@namedef{PYG@tok@nf}{\def\PYG@tc##1{\textcolor[rgb]{0.00,0.00,1.00}{##1}}}
\@namedef{PYG@tok@nc}{\let\PYG@bf=\textbf\def\PYG@tc##1{\textcolor[rgb]{0.00,0.00,1.00}{##1}}}
\@namedef{PYG@tok@nn}{\let\PYG@bf=\textbf\def\PYG@tc##1{\textcolor[rgb]{0.00,0.00,1.00}{##1}}}
\@namedef{PYG@tok@ne}{\let\PYG@bf=\textbf\def\PYG@tc##1{\textcolor[rgb]{0.80,0.25,0.22}{##1}}}
\@namedef{PYG@tok@nv}{\def\PYG@tc##1{\textcolor[rgb]{0.10,0.09,0.49}{##1}}}
\@namedef{PYG@tok@no}{\def\PYG@tc##1{\textcolor[rgb]{0.53,0.00,0.00}{##1}}}
\@namedef{PYG@tok@nl}{\def\PYG@tc##1{\textcolor[rgb]{0.46,0.46,0.00}{##1}}}
\@namedef{PYG@tok@ni}{\let\PYG@bf=\textbf\def\PYG@tc##1{\textcolor[rgb]{0.44,0.44,0.44}{##1}}}
\@namedef{PYG@tok@na}{\def\PYG@tc##1{\textcolor[rgb]{0.41,0.47,0.13}{##1}}}
\@namedef{PYG@tok@nt}{\let\PYG@bf=\textbf\def\PYG@tc##1{\textcolor[rgb]{0.00,0.50,0.00}{##1}}}
\@namedef{PYG@tok@nd}{\def\PYG@tc##1{\textcolor[rgb]{0.67,0.13,1.00}{##1}}}
\@namedef{PYG@tok@s}{\def\PYG@tc##1{\textcolor[rgb]{0.73,0.13,0.13}{##1}}}
\@namedef{PYG@tok@sd}{\let\PYG@it=\textit\def\PYG@tc##1{\textcolor[rgb]{0.73,0.13,0.13}{##1}}}
\@namedef{PYG@tok@si}{\let\PYG@bf=\textbf\def\PYG@tc##1{\textcolor[rgb]{0.64,0.35,0.47}{##1}}}
\@namedef{PYG@tok@se}{\let\PYG@bf=\textbf\def\PYG@tc##1{\textcolor[rgb]{0.67,0.36,0.12}{##1}}}
\@namedef{PYG@tok@sr}{\def\PYG@tc##1{\textcolor[rgb]{0.64,0.35,0.47}{##1}}}
\@namedef{PYG@tok@ss}{\def\PYG@tc##1{\textcolor[rgb]{0.10,0.09,0.49}{##1}}}
\@namedef{PYG@tok@sx}{\def\PYG@tc##1{\textcolor[rgb]{0.00,0.50,0.00}{##1}}}
\@namedef{PYG@tok@m}{\def\PYG@tc##1{\textcolor[rgb]{0.40,0.40,0.40}{##1}}}
\@namedef{PYG@tok@gh}{\let\PYG@bf=\textbf\def\PYG@tc##1{\textcolor[rgb]{0.00,0.00,0.50}{##1}}}
\@namedef{PYG@tok@gu}{\let\PYG@bf=\textbf\def\PYG@tc##1{\textcolor[rgb]{0.50,0.00,0.50}{##1}}}
\@namedef{PYG@tok@gd}{\def\PYG@tc##1{\textcolor[rgb]{0.63,0.00,0.00}{##1}}}
\@namedef{PYG@tok@gi}{\def\PYG@tc##1{\textcolor[rgb]{0.00,0.52,0.00}{##1}}}
\@namedef{PYG@tok@gr}{\def\PYG@tc##1{\textcolor[rgb]{0.89,0.00,0.00}{##1}}}
\@namedef{PYG@tok@ge}{\let\PYG@it=\textit}
\@namedef{PYG@tok@gs}{\let\PYG@bf=\textbf}
\@namedef{PYG@tok@gp}{\let\PYG@bf=\textbf\def\PYG@tc##1{\textcolor[rgb]{0.00,0.00,0.50}{##1}}}
\@namedef{PYG@tok@go}{\def\PYG@tc##1{\textcolor[rgb]{0.44,0.44,0.44}{##1}}}
\@namedef{PYG@tok@gt}{\def\PYG@tc##1{\textcolor[rgb]{0.00,0.27,0.87}{##1}}}
\@namedef{PYG@tok@err}{\def\PYG@bc##1{{\setlength{\fboxsep}{\string -\fboxrule}\fcolorbox[rgb]{1.00,0.00,0.00}{1,1,1}{\strut ##1}}}}
\@namedef{PYG@tok@kc}{\let\PYG@bf=\textbf\def\PYG@tc##1{\textcolor[rgb]{0.00,0.50,0.00}{##1}}}
\@namedef{PYG@tok@kd}{\let\PYG@bf=\textbf\def\PYG@tc##1{\textcolor[rgb]{0.00,0.50,0.00}{##1}}}
\@namedef{PYG@tok@kn}{\let\PYG@bf=\textbf\def\PYG@tc##1{\textcolor[rgb]{0.00,0.50,0.00}{##1}}}
\@namedef{PYG@tok@kr}{\let\PYG@bf=\textbf\def\PYG@tc##1{\textcolor[rgb]{0.00,0.50,0.00}{##1}}}
\@namedef{PYG@tok@bp}{\def\PYG@tc##1{\textcolor[rgb]{0.00,0.50,0.00}{##1}}}
\@namedef{PYG@tok@fm}{\def\PYG@tc##1{\textcolor[rgb]{0.00,0.00,1.00}{##1}}}
\@namedef{PYG@tok@vc}{\def\PYG@tc##1{\textcolor[rgb]{0.10,0.09,0.49}{##1}}}
\@namedef{PYG@tok@vg}{\def\PYG@tc##1{\textcolor[rgb]{0.10,0.09,0.49}{##1}}}
\@namedef{PYG@tok@vi}{\def\PYG@tc##1{\textcolor[rgb]{0.10,0.09,0.49}{##1}}}
\@namedef{PYG@tok@vm}{\def\PYG@tc##1{\textcolor[rgb]{0.10,0.09,0.49}{##1}}}
\@namedef{PYG@tok@sa}{\def\PYG@tc##1{\textcolor[rgb]{0.73,0.13,0.13}{##1}}}
\@namedef{PYG@tok@sb}{\def\PYG@tc##1{\textcolor[rgb]{0.73,0.13,0.13}{##1}}}
\@namedef{PYG@tok@sc}{\def\PYG@tc##1{\textcolor[rgb]{0.73,0.13,0.13}{##1}}}
\@namedef{PYG@tok@dl}{\def\PYG@tc##1{\textcolor[rgb]{0.73,0.13,0.13}{##1}}}
\@namedef{PYG@tok@s2}{\def\PYG@tc##1{\textcolor[rgb]{0.73,0.13,0.13}{##1}}}
\@namedef{PYG@tok@sh}{\def\PYG@tc##1{\textcolor[rgb]{0.73,0.13,0.13}{##1}}}
\@namedef{PYG@tok@s1}{\def\PYG@tc##1{\textcolor[rgb]{0.73,0.13,0.13}{##1}}}
\@namedef{PYG@tok@mb}{\def\PYG@tc##1{\textcolor[rgb]{0.40,0.40,0.40}{##1}}}
\@namedef{PYG@tok@mf}{\def\PYG@tc##1{\textcolor[rgb]{0.40,0.40,0.40}{##1}}}
\@namedef{PYG@tok@mh}{\def\PYG@tc##1{\textcolor[rgb]{0.40,0.40,0.40}{##1}}}
\@namedef{PYG@tok@mi}{\def\PYG@tc##1{\textcolor[rgb]{0.40,0.40,0.40}{##1}}}
\@namedef{PYG@tok@il}{\def\PYG@tc##1{\textcolor[rgb]{0.40,0.40,0.40}{##1}}}
\@namedef{PYG@tok@mo}{\def\PYG@tc##1{\textcolor[rgb]{0.40,0.40,0.40}{##1}}}
\@namedef{PYG@tok@ch}{\let\PYG@it=\textit\def\PYG@tc##1{\textcolor[rgb]{0.24,0.48,0.48}{##1}}}
\@namedef{PYG@tok@cm}{\let\PYG@it=\textit\def\PYG@tc##1{\textcolor[rgb]{0.24,0.48,0.48}{##1}}}
\@namedef{PYG@tok@cpf}{\let\PYG@it=\textit\def\PYG@tc##1{\textcolor[rgb]{0.24,0.48,0.48}{##1}}}
\@namedef{PYG@tok@c1}{\let\PYG@it=\textit\def\PYG@tc##1{\textcolor[rgb]{0.24,0.48,0.48}{##1}}}
\@namedef{PYG@tok@cs}{\let\PYG@it=\textit\def\PYG@tc##1{\textcolor[rgb]{0.24,0.48,0.48}{##1}}}

\def\PYGZus{\char`\_}
\def\PYGZob{\char`\{}
\def\PYGZcb{\char`\}}

\def\PYGZam{\char`\&}
\def\PYGZlt{\char`\<}
\def\PYGZgt{\char`\>}
\def\PYGZsh{\char`\#}

\def\PYGZhy{\char`\-}
\def\PYGZsq{\char`\'}
\def\PYGZdq{\char`\"}

\makeatother

\makeatletter
\Crefname{algorithm}{Algo.}{Algorithms}
\Crefname{table}{Tab.}{Tables}
\crefname{section}{\S\@gobble}{\S\S\@gobble}
\crefname{subsection}{\S\@gobble}{\S\S\@gobble}
\makeatother

\newcommand{\eg}{\hbox{\emph{e.g.}}\xspace}
\newcommand{\ie}{\hbox{\emph{i.e.}}\xspace}

\newcommand\code{\ensuremath{\bm{y}}}

\newcommand\intent{\ensuremath{\bm{x}}}
\newcommand\intentset{\ensuremath{\bm{X}}}

\newcommand\context{\ensuremath{\bm{c}}}
\newcommand\contextset{\ensuremath{\bm{C}}}
\newcommand\spec{\ensuremath{\bm{z}}}
\newcommand{\var}{\ensuremath{{v}}}
\newcommand{\promptset}{\ensuremath{\mathcal{P}}}

\newcommand\utterance[1]{\textit{#1}}

\newcommand{\iospecbg}[1]{\tikz[baseline=-0.5ex] \node [fill=iospeccolor,rounded corners,font=\normalsize,inner sep=1.5pt] at (0,0) {#1};}

\newcommand{\intentbg}[1]{\tikz[baseline=-0.5ex] \node [fill=intentcolor,rounded corners,font=\normalsize,inner sep=1.5pt] at (0,0) {#1};}

\usepackage{ragged2e}

\DeclareTextFontCommand{\emph}{\em}

\usepackage{amsmath,amsfonts,bm}
\usepackage{bbm}
\usepackage[nameinlink]{cleveref}
\creflabelformat{equation}{#2\textup{#1}#3}  %
\Crefname{equation}{Eq.}{Eqs.}
\Crefname{figure}{Fig.}{Figs.}
\Crefname{table}{Table}{Tables}

\def\eqref#1{equation~\ref{#1}}

\def\1{\bm{1}}

\DeclareMathAlphabet{\mathsfit}{\encodingdefault}{\sfdefault}{m}{sl}
\SetMathAlphabet{\mathsfit}{bold}{\encodingdefault}{\sfdefault}{bx}{n}

\newcommand{\N}{\mathbb{N}}

\def\nbmodel/{code LLM}
\def\palmsixtytwob/{\textsc{PaLM 62B}}
\def\palm/{\textsc{PaLM}}
\def\palm2/{\textsc{PaLM2}}
\def\giftforcode/{\textsc{Gift4Code}}

\newcommand{\SyntheticDataset}{\mathcal{D}}

\newcommand{\passat}[1]{\textit{pass}@\ensuremath{#1}}
\newcommand{\bigpassat}[1]{\textit{Pass}@\ensuremath{#1}}

\def\arcade/{\textsc{Arcade}}
\def\dsonek/{\textsc{DS-1000}}

\lstdefinestyle{program}{
  basicstyle=\fontfamily{cmtt}\small,
  language=Python,
  otherkeywords={self,call,append,split,write},  %
  keywordstyle=\bfseries\color{deepblue},
  emph={},          %
  emphstyle=\color{deepred},    %
  showstringspaces=false,
  breaklines=true,
  columns=fullflexible,
}

\usepackage[textsize=tiny]{todonotes}

\renewcommand{\paragraph}[1]{\noindent \textbf{#1}}

\icmltitlerunning{Grounding Data Science Code Generation with Input-Output Specifications}

\begin{document}

\twocolumn[
\icmltitle{
Grounding Data Science Code Generation with Input-Output Specifications
}

\icmlsetsymbol{intern}{*}

\begin{icmlauthorlist}
\icmlauthor{Yeming Wen}{yyy}
\icmlauthor{Pengcheng Yin}{comp}
\icmlauthor{Kensen Shi}{comp}
\icmlauthor{Henryk Michalewski}{comp}
\icmlauthor{Swarat Chaudhuri}{yyy}
\icmlauthor{Alex Polozov}{comp}
\end{icmlauthorlist}

\icmlaffiliation{yyy}{Department of Computer Science, University of Texas at Austin}
\icmlaffiliation{comp}{Google Inc.}

\icmlcorrespondingauthor{Yeming Wen, Work done during Internship at Google.}{ywen@utexas.edu}

\icmlkeywords{Machine Learning, ICML}

\vskip 0.3in
]

\preprintAffiliationsAndNotice{}

\begin{abstract}
Large language models (LLMs) have recently demonstrated a remarkable ability to generate code from natural language (NL) prompts. However, in the real world, NL is often too ambiguous to capture the true intent behind programming problems, requiring additional input-output (I/O) specifications.
Unfortunately, LLMs can have difficulty aligning their outputs with both the NL prompt and the I/O specification.
In this paper, we give a way to mitigate this issue in the context of data science programming, where tasks require explicit I/O specifications for clarity.
Specifically, we propose \giftforcode/, a novel approach for the instruction fine-tuning of LLMs with respect to I/O specifications.
Our method leverages synthetic data produced by the LLM itself and utilizes execution-derived feedback as a key learning signal.
This feedback, in the form of program I/O specifications, is provided to the LLM to facilitate instruction fine-tuning. 
We evaluated our approach on two challenging data science benchmarks, \arcade/ and \dsonek/.
The results demonstrate a significant improvement in the LLM's ability to generate code that is not only executable but also accurately aligned with user specifications, substantially improving the quality of code generation for complex data science tasks.
\end{abstract}

\section{Introduction}

\label{sec:intro}

\begin{figure*}[t]
    \centering
    \includegraphics[width=0.9\textwidth]{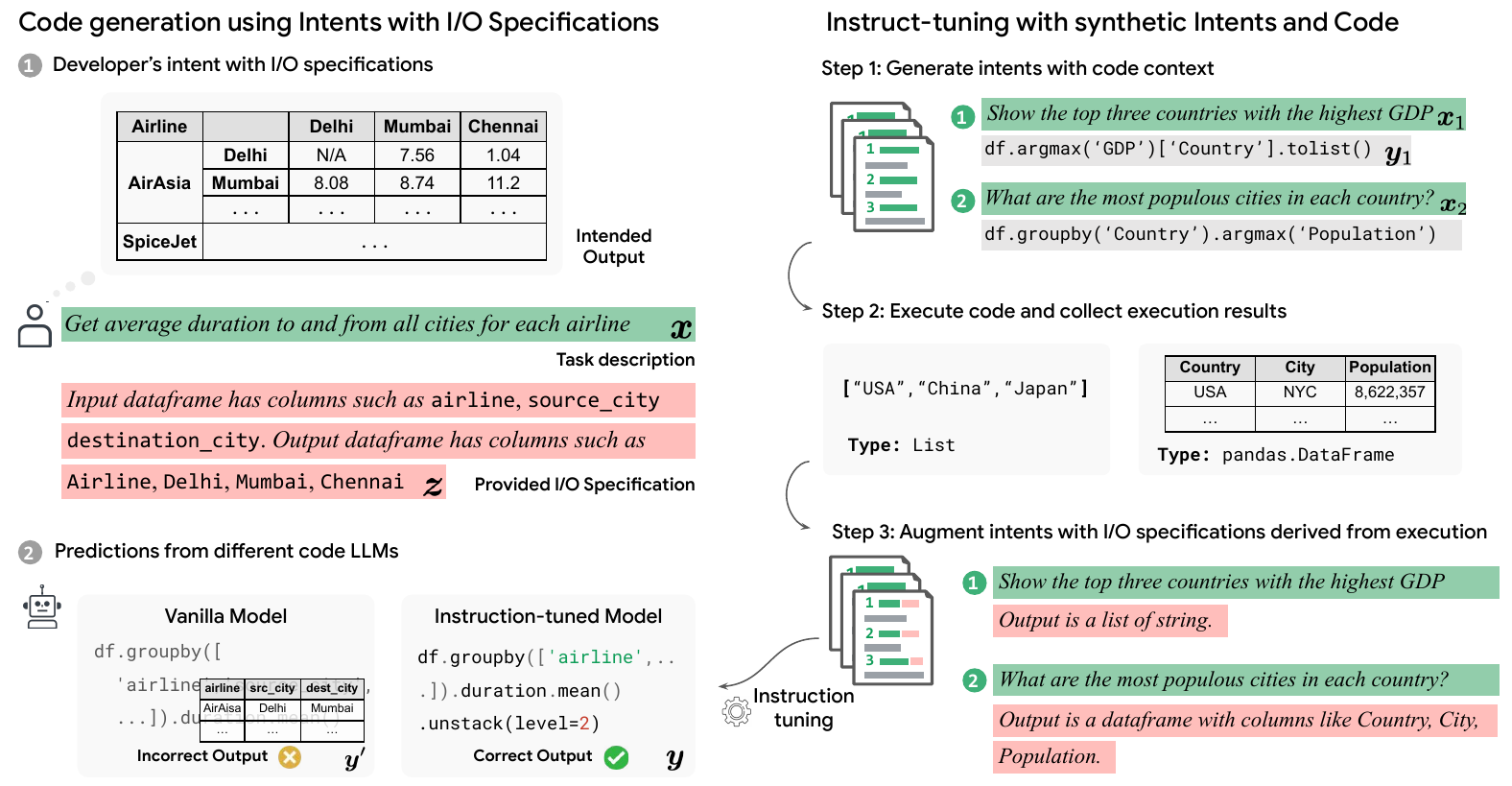}
    \caption{\textbf{\textit{Left}:} Illustration of how users prompt code LLMs with NL intents and I/O specifications to generate code with complex outputs (\texttt{pandas} \texttt{Dataframe}s). Vanilla code LLMs fail to understand I/O specifications.
    \textbf{\textit{Right}:} Our proposed instruction tuning approach uses synthetic intents and code solutions, where intents are augmented with I/O specifications derived from program execution. Models trained on such synthetic data could better follow a user's intent.}
    \label{fig:intro_teaser}
    \vspace{-1em}
\end{figure*}

Large language models (LLMs) have recently shown tremendous promise at generating code from natural language prompts~\citep{Chen2021EvaluatingLL,austin2021lambdacode,Li2023StarCoderMT,Li2022CompetitionlevelCG,Nijkamp2022CodeGenAO,fried2022incoder,Li2023StarCoderMT}.
In particular, LLMs trained on code have been shown to excel at solving interview-style coding problems, represented by benchmarks such as HumanEval
~\citep{Chen2021EvaluatingLL}. In these tasks, natural language (NL) ``intents" (\eg~$\intent$ in \cref{fig:intro_teaser})  can succinctly define the programming problem at hand. The objective of the LLM is to produce output that is aligned with this intent.

However, in the real world, not all programming tasks 
can be easily described using compact NL intents. 
Among these, data science programming stands out as a vital domain, where data scientists spend considerable time on data wrangling and exploratory data analysis (EDA).
Whereas problems in benchmarks like HumanEval involve basic Python types that can be easily tested, 
data science tasks often involve complex operations on data structures like \texttt{Pandas} DataFrames or \texttt{PyTorch} Tensors.
The inherent ambiguity in expressing these tasks through NL intents (\eg~\intentbg{green text} in \cref{fig:intro_teaser}) often leads to misalignment between developers' requirements and the code generated~\citep{yin22acl}. 

Input-output (I/O) specifications (\eg~\iospecbg{red text} in \cref{fig:intro_teaser}), ranging from concrete I/O examples to high-level NL summaries, are a natural way to reduce the  ambiguity of NL intents~\citep{gulwani2015inductive,Balog2016DeepCoderLT,jain2022jigsaw}.
Prior to the emergence of LLMs, specifications served as essential problem descriptions in program synthesis~\citep{Gulwani2016ProgrammingBE,Devlin2017RobustFillNP,Shi2020TFCoderPS}.
Real-world synthesis systems like FlashFill are testimony to the adoption and effectiveness of I/O specifications~\citep{Gulwani2011AutomatingSP,Gulwani2012SpreadsheetDM}.
In this work, we consider the problem of LLM-based code generation in data science when the prompt provided to the LLM consists of both a NL intent and an additional I/O specification.

Unfortunately, code LLMs often fail to follow intents with additional semantic constraints like I/O specifications out-of-the-box, leading to plausible solutions that fail to satisfy the provided specifications (\eg~$\code'$, \cref{fig:intro_teaser}).
This can pose unnecessary burden on developers who are then required to fix the generated code~\citep{bird23flight}.
Such a misalignment between the user's intent and the model's predictions primarily stems from the \emph{lack} of training data formatted with such specifications.

Instruction fine-tuning has emerged as an effective strategy to tackle the issue of misalignment~\citep{Wei2021FinetunedLM,Sanh2021MultitaskPT,Chung2022ScalingIL}. 
Classical approaches for instruction tuning typically require a substantial amount of parallel labeled data of NL intents and gold model responses.
The process of gathering such data is labor-intensive and time-consuming. 
Recent studies have suggested that generating synthetic instruction-following data using the LLM itself is a promising approach to improve alignment, with empirical success on natural language text generation tasks~\citep[][\textit{inter alia}]{Wang2022SelfInstructAL,Honovich2022UnnaturalIT,alpaca,Peng2023InstructionTW}.

In this paper we build upon the recent success of instruction tuning using synthetic data and fine-tune code LLMs to follow NL intents with additional I/O specifications.
Unlike existing approaches, our key insight is to leverage program execution for synthetic data generation.
First, in contrast to other open-ended text generation tasks where assessing the quality of target responses is challenging, the quality of synthetic code generation data can be easily improved using heuristics such as code executability~\citep{yin22acl}.
Moreover, from the program execution states one could derive precise and \emph{aligned} I/O specifications that can be included in the intents to supervise a model to follow those extra semantic constraints (\cref{fig:intro_teaser}, \textit{Right}).
In other words, when fine-tuned on such synthetic data, a model learns to \emph{ground} NL \intentbg{task descriptions} to program execution states expressed as \iospecbg{I/O specifications}~\citep{berant2013freebase}.

We apply our \underline{g}rounded \underline{i}nstruction \underline{f}ine-\underline{t}uning for \underline{c}ode (\giftforcode/) method to two challenging natural language to code generation applications: synthesizing complex \texttt{pandas} programs in computational notebooks (\arcade/, \citet{yin22acl}) and answering data science questions on Stack Overflow (\dsonek/, \citet{Lai2022DS1000}).
First, we demonstrate the value of leveraging program execution information by showing that strong code LLMs can still be improved after fine-tuning on our synthetic data \emph{without} including any I/O specifications. 
Then, to further align model predictions to various types of user-provided I/O specifications, we derive those specifications at different levels of abstraction from code execution results.
This ranges from concrete input/output examples to NL summaries of target variables (\iospecbg{specifications} in \cref{fig:intro_teaser}).
By fine-tuning on parallel data of intents with I/O specifications and their target code solutions, the model is better at producing code that is more likely to execute to the desired outcome.
Although our experiments focused on the data science domain, it's important to note that the methodology underlying \giftforcode/ is general and adaptable. It can be  applied to different domains that require specifications for a precise task description.

\section{Problem Formulation}

\paragraph{Natural Language to Code Generation}
Code generation considers the task of translating a developer's natural language intent $\intent$ into a machine-executable program $\code$ (\eg~\cref{fig:intro_teaser}, \textit{Left}).
An intent usually contains a succinct and potentially ambiguous \intentbg{task description}.
For tasks with complex outputs, the intent may also include additional \iospecbg{I/O specifications} as extra clarifications.\footnote{For simplicity, we consider I/O specifications as part of the intent hereafter.}
Code generation tasks are often contextualized, meaning that an intent is associated with certain \emph{programmatic contexts} $\context$~\citep{Iyer2018MappingLT}, such as the code that a developer has already written in an IDE prior to the intent (\eg~\texttt{df\hspace{0.5mm}=\hspace{0.5mm}pd.read\_csv(\textquotedbl flights.csv\textquotedbl)} for \intent, not shown in \cref{fig:intro_teaser}).
Intuitively, a model needs to leverage both the intent and the programmatic context (\eg~variable \texttt{df}) to generate a suitable code solution.

\paragraph{Supervised Instruction Tuning for Code LLMs}
Supervised instruction tuning aims to improve code LLMs by fine-tuning them on parallel data of intents and target code solutions.
In this paper we consider automatically synthesizing such parallel data by prompting LLMs using few-shot demonstrations (other approaches are discussed in \cref{sec:related_work}).
Specifically, the synthetic dataset consists of examples $\{ \langle \context, \intent, \code \rangle \}$ of intents $\intent$ with programmatic contexts $\context$ and their generated code solutions $\code$.
The pivotal question then becomes how to create a high-quality synthetic $\{ \langle \context, \intent, \code \rangle \}$ that facilitates the instruction fine-tuning process.

\section{\giftforcode/: Learning to Follow Intents with I/O Specifications}
\label{sec:GIFT4Code}

In this section we elaborate on \giftforcode/, our proposed approach to fine-tune code LLMs to better follow developers' natural language intents along with I/O specifications, using synthetic parallel data. \cref{fig:intro_teaser}(\textit{Right}) illustrates an overview of \giftforcode/.
We first synthesize a collection of intents with code solutions via few-shot prompting (\cref{sec:method:synthetic_data_generation}), and then execute model-predicted code to derive I/O specifications from execution results (\cref{subsec:code_execution_and_io_inference}).
Finally, we fine-tune the code LLM to predict code solutions given intents inlined with I/O specifications (\cref{sec:method:instruction_tuning}).

\subsection{Generating Synthetic Intents and Code Solutions}
\label{sec:method:synthetic_data_generation}

\paragraph{Programmatic Contexts}
We initialize a program state given some programmatic context 
and generate a series of contextualized NL-to-code problems for that context.
As an example, the synthetic problems in \cref{fig:intro_teaser} (\textit{Right}) could have the contextual code \texttt{df = pd.read\_csv(\textquotedbl world\_statistics.csv\textquotedbl)}, which initializes the DataFrame variable \texttt{df}, subsequently used in the generated synthetic examples.
The fact that our problems are contextualized sets our approach apart from existing instruction-tuning methods for text generation models~\citep{Wang2022SelfInstructAL,Honovich2022UnnaturalIT}, where synthetic examples do not depend on any particular contexts.
In our case, we mine those programmatic contexts from real-world code repositories, such as tabular datasets (e.g., \texttt{.csv}) used in data science notebooks on Github (\cref{sec:experiments}).

\paragraph{Creating Initial NL Intents}
Given a programmatic context \context, we few-shot prompt an LLM to create a sequence of \intentbg{natural language intents} $\{ \intent_i \}$ (\eg~$\intent_1$, $\intent_2$ in \cref{fig:intro_teaser}(\textit{Right})).
A problem $\intent_i$ that appears later in the sequence might depend on the earlier ones $\{ \intent_{<i} \}$~\citep{Nijkamp2022CodeGenAO,yin22acl}.
To generate NL intents, we use a \enquote{generalist} \palm2/ LLM~\citep{palm2}, instead of the code LLM that we aim to improve, since predicting intents conditioned on some context is similar to other text generation tasks, which could be better handled by a LM trained on general-purpose text data~\citep{Zelikman2022Parsel}. 
The \palm2/ LLM achieves competitive results with GPT-4 on a variety of NL reasoning tasks.
Empirically, we observe that the problems generated by this LLM cover a wide range of tasks relevant to the given programmatic context. Readers can refer to \cref{sec:appendix_llm_generated_nl_intents} for examples. Note that those model-predicted intents do not come with I/O specifications yet.

\paragraph{Predicting Code Solutions}
After generating an intent \intent, we then prompt the code LLM to get a code solution $\code$ for $\intent$ (\eg~$\code_1$ in \cref{fig:intro_teaser}(\textit{Right})).
Specifically, a prompt to the LLM is the concatenation of the programmatic context $\context$ and the intent \intent, with additional few-shot demonstrations of $\{ \langle \context', \intent', \code' \rangle \}$.
Since many NL intents can be ambiguous and there could exist multiple alternative solutions (\eg~without additional I/O specifications, the intent in green in \cref{fig:intro_teaser}(\textit{Left}) could be answered using tables with different layouts; see more in \cite{yin22acl}), we therefore draw multiple candidate code solutions $\{ \code \}$ for each intent.
Intuitively, $\{ \code \}$ could have a variety of alternative solutions for $\intent$, each leading to different execution results. This equips the model with the capacity to predict code for the same task but with different user-provided I/O specifications.

\paragraph{Improving Quality of Synthetic Data}
The quality of synthetic data is a fundamental issue of data augmentation in instruction tuning~\citep{Wang2022SelfInstructAL,Honovich2022UnnaturalIT}, and existing approaches in text generation typically resort to simple and noisy heuristics (\eg~rejecting examples with different inputs but the same output).
As motivated in \cref{sec:intro}, for NL-to-code generation, we can reliably enhance the quality of candidate code solutions by leveraging inherent program properties, such as filtering out any code that is not executable given the provided programmatic context.

\def\typedesc/{$\mathsf{Type Desc}$}
\def\ioexample/{$\mathsf{I/O~Examples}$}
\def\iosummary/{$\mathsf{I/O~Summary}$}

\begin{table*}[t]
    \small
    \centering
    \setlength{\tabcolsep}{5pt}
    \begin{tabular}{lll}
    \toprule
    \textbf{Spec.~Type} & \textbf{Description} & \textbf{Example I/O Specification} \\ \midrule
    \typedesc/ & \begin{minipage}{0.25 \textwidth}Variable type name\end{minipage} & \begin{minipage}{0.55 \textwidth}Generate a variable with name \texttt{df} and type \texttt{pandas.DataFrame}\end{minipage} \\ \midrule
    \ioexample/ & \begin{minipage}{0.25 \textwidth}Concrete I/O examples \end{minipage} & 
    \begin{minipage}{0.55 \textwidth}
        Output variable \texttt{df}: 
        \begin{lstlisting}[style=program,basicstyle=\fontfamily{cmtt}\scriptsize]
|Bangalore(float)|Chennai(float)|Delhi(float)|Hyderabad
(float)|Kolkata(float)|Hyderabad(float)|Kolkata(float)|...
|-----|-----|-----|-----|-----|-----|
| nan | 1.04 | 8.08 | 3.62 | 7.56 | 7.56 | 8.32 |
| 1.18 | nan | 11.96 | 6.80 | 6.31 | 8.75 |
| 8.46 | 11.10 | nan | 9.19 | 9.52 | 10.32 |
    \end{lstlisting}
    \end{minipage}
    \\ \midrule
    \iosummary/ & \begin{minipage}{0.25 \textwidth}LLM-generated NL summaries of I/O examples\end{minipage} & \begin{minipage}{0.55 \textwidth}Given the user intent and the code, the salient columns (at most given 3) in the input dataframe are airline, source\_city, destination\_city. The output dataframe has columns (at most given 3) such as Delhi, Mumbai, Chennai.\end{minipage} \\ \bottomrule
    \end{tabular}
    \caption{Types of I/O specifications proposed in this work at different levels of abstraction. Example specifications are for the intent in \cref{fig:intro_teaser}(\textit{Left}). Only the output specifications for \ioexample/ are shown for brevity.}
    \label{tab:output_spec:examples}
    \vspace{-0.5em}
\end{table*}

\subsection{Code Execution \& Inference of I/O Specifications}
\label{subsec:code_execution_and_io_inference}

Given the set of synthetic problems $\{ \langle \intent, \{ \code \} \rangle \}$ generated by few-shot prompting, we execute the code for each problem (step 2, \cref{fig:intro_teaser}(\textit{Right})) and derive \iospecbg{I/O specifications} from the execution results as additional semantic constraints to be included in the intents (step 3, \cref{fig:intro_teaser}(\textit{Right})).

Specifically, for each candidate solution $\code$ of an intent, we first execute its original programmatic context $\context$, followed by executing $\code$.
We trace the execution to collect the set of input and output variables in $\code$, denoted as $\{ \var \}$, which are used to derive I/O specifications (details below).
Executing code with arbitrary programmatic contexts collected from the wild is highly non-trivial due to issues such as library dependency. However, the use of synthetic data alleviates the need for a complex environment setup.

Given the set of input and output variables extracted from execution results, we formulate an I/O specification, denoted as $\spec$, which serves as additional information to augment a developer's intent, thereby providing a more comprehensive problem description.
The level of detail and the style of these I/O specifications can vary based on the complexity of the problem and the developer's preferences. In this work, we investigate three distinct types of I/O specifications, each characterized by its own linguistic style and level of abstraction, as illustrated in \cref{tab:output_spec:examples}. 

First, as a simple baseline, we utilize the variable type ({\typedesc/}, \cref{tab:output_spec:examples}) as the I/O specification. Next, we incorporate the concrete values of the input/output variables into the specification, which we refer to as {\ioexample/}.
This is reminiscent of classical program synthesis using I/O examples~\citep{Gulwani2012SpreadsheetDM,Alur2013SyntaxguidedS,Balog2016DeepCoderLT}.
However, in our scenario, these I/O examples are used in conjunction with natural language (NL) intents to define the problem, in line with~\cite{jain2022jigsaw}.
Given that the majority of the problems in our synthetic dataset involve complex Python objects like \texttt{pandas} \texttt{DataFrame}s, we simplify the I/O specification to include only partial variable states (\eg~by excluding some rows and columns in large \texttt{DataFrame}s). Please refer to \cref{sec:appendix_iosummary_examples} for details.

In our effort to generate a more natural variety of I/O specifications that closely resemble the style of specifications in developers' NL intents,
we employ an LLM to summarize the values of input/output variables $\{ \var \}$ into a succinct natural language description $\spec$ (\iosummary/).
Intuitively, the NL I/O summary includes salient information in the variables that can best clarify the original intent (\eg~the subset of columns in a \texttt{DataFrame} that are most relevant to solve a problem, as in \cref{tab:output_spec:examples}, \textit{Bottom}).

Specifically, we few-shot prompt the \enquote{generalist} \palm2/ to generate $\spec$, using information from its programmatic context \context, the intent \intent, the code solution $\code$, as well as I/O variables $\{ \var \}$, \ie~$\spec \sim P_{\textrm{LLM}}( \cdot\ |\ \context, \intent, \code, \{ \var \} )$.
We then update the intent $\intent$ by appending $\spec$ to it.
The few-shot exemplars used for prompting cover example I/O summaries for various types of Python objects, such as nested container types (\eg~nested \texttt{dict}s), along with more complex objects like \texttt{Pandas} \texttt{DataFrame}s and \texttt{PyTorch} or \texttt{Tensorflow} \texttt{Tensor}s. See \cref{sec:appendix_iosummary_examples} for additional details.

\subsection{Fine-tuning Code LLMs to Follow Intents with I/O Specifications}
\label{sec:method:instruction_tuning}

Our approach, \giftforcode/, aims to fine-tune code LLMs to generate code that adheres closely to the desired intents which are supplemented by I/O specifications.
In our synthetic training data, each example $\langle \context, \intent, \code \rangle$ consists of a programmatic context $\context$, an intent $\intent$ augmented with I/O specifications, and the corresponding code solution $\code$.
During fine-tuning, the code LLM learns to generate code that not only satisfies the provided intents but also respects the specified I/O constraints, while leveraging any relevant information in the programmatic contexts.
In other words, we optimize $P_\textrm{LLM}(\code\ |\ \context, \intent)$. It is worth noting that the code LLM that undergoes this optimization is different from the \enquote{generalist} \palm2/ LLM employed to generate the NL intents and I/O specification $\spec$.
\footnote{The \enquote{generalist} LLM, such as \palm2/ or GPT-4, is typically accessible only through an API.}

\section{Experiments}
\label{sec:experiments}

The core research question explored in this section is whether \giftforcode/ enhances the LLM's ability to follow developers’ NL intents with \emph{\textbf{complex}} I/O specifications. 
While common code generation benchmarks like HumanEval~\citep{Chen2021EvaluatingLL} and MBPP~\citep{austin2021lambdacode} feature simple algorithmic tasks (e.g., sorting) utilizing basic Python data types (e.g., lists), thus allowing for the use of concrete I/O examples as specifications, they lack the diverse and complex I/O specifications that we aim to explore. 
For more open-ended tasks in the data science domain, the output data type is more complex and diverse (e.g., \texttt{pandas} \texttt{DataFrame}s, \texttt{pytorch} \texttt{Tensor}s ).
Hence, we applied our method to two different data science code generation applications.

\noindent \textbf{\arcade/}~\citep{yin22acl} is a benchmark of natural language to code generation in interactive data science notebooks.
Each evaluation notebook consists of a series of interrelated NL-to-code problems in data wrangling (\eg~\utterance{``Min-max normalize numeric columns''}) and exploratory data analysis (\eg~intents in \cref{fig:intro_teaser}) using the \texttt{pandas} library.
\arcade/ features succinct NL intents to reflect the style of ephemeral queries from developers when prompting LLMs for code completion. 
More than 50\% of the dataset's problems are under-specified, which means that additional I/O specifications could provide extra clarification.
To construct programmatic contexts for synthetic training data generation, we scraped 7,500 CSV files that are used in public Jupyter notebooks. 
Each context contains a DataFrame import statement, for example, \texttt{df = pd.read\_csv($\cdot$)}, followed by an NL description of the DataFrame to help the LLM understand its content.
We generated 6 intents for each programmatic context and sampled 5 candidate code solutions for each intent.
Roughly 60\% of the code samples were executable.
After filtering based on executability and API diversity (\cref{sec:method:synthetic_data_generation}), we obtained around $~$20$K$ synthetic examples for instruction fine-tuning.

Our synthetic data only comprises pairs of questions and code samples which lack rich context. 
To avoid regression in context understanding during instruction fine-tuning, we crafted a mixture dataset which combines the synthetic data and the Python data used to fine-tune the \nbmodel/. Note that this Python data does not contain any execution-based signals or I/O specifications. After approximately 1,500 instruction tuning steps with a batch size of 64, the model reaches its optimal performance. This process consumed about 1.5 epochs of our synthetic dataset.

\noindent \textbf{\dsonek/} \citep{Lai2022DS1000} is a benchmark of data science problems sourced from Stack Overflow (\textsc{SO}).
Compared to \arcade/, problems in \dsonek/ feature a wider variety of I/O types, such as \texttt{numpy}/\texttt{scipy} \texttt{Array}s and \texttt{pytorch}/\texttt{tensorflow}  \texttt{Tensors}, making it particularly appealing to evaluate our instruction tuning approach aimed at generating code following I/O specifications.
However, in contrast to \arcade/ which features succinct NL intents, \dsonek/ follows the typical style of detailed problem descriptions found in SO posts. These elaborate descriptions often include additional information such as task background and descriptions of unsuccessful attempts with an average length of 140 words.
Given that such elaborate intents may not reflect the style of developers' prompts to code LLMs, we do not focus on generating intents with similar styles.
Instead, we held-out 500 problems in \dsonek/ and use their annotated intents as training data, while evaluating on the remaining problems.\footnote{We only use the annotated intents, while the code solutions and I/O specifications are still predicted by the LLM. We ensure the training and evaluation problems are disjoint and from different Stack Overflow posts.}

\begin{table*}[t]
    \small
    \centering
    \def\arraystretch{1}
    \setlength{\tabcolsep}{10pt}
    \begin{tabular}{@{}clcccc:c@{}}
    \toprule
     \multirow{3}{*}{\textbf{Methods}} & 
     \multirow{3}{*}{\quad \medspace \textbf{Spec. Type}} & 
     \multicolumn{4}{c}{\textbf{\arcade/}} & \textbf{\dsonek/} \\

     & & \multicolumn{2}{c}{\passat{5}} & \multicolumn{2}{c}{\passat{20}} & \multirow{2}{*}{\passat{1}} \\
     
      \cmidrule(lr){3-4} \cmidrule(lr){5-6}
      & & \textrm{No Context} & \textrm{Full Context} & \textrm{No Context} & \textrm{Full Context} \\ 
    \rowcolor{black!10!} \multicolumn{7}{c}{\textbf{\textit{Zero-shot Prompting}}} \\
    StarCoder 15B & \qquad \medspace no spec. & 11.75 & 22.38 & 17.24 & 32.52 & 21.52 \\
    \midrule
    Code LLM 62B &  \qquad \medspace no spec. & 12.45 &  27.75 & 19.85 & 37.47 & 22.62 \\
    ~~$+$ \giftforcode/ & \qquad \medspace no spec. & 20.78 & 34.33 & 33.40 & 46.94 & 24.56 \\
    \rowcolor{black!10!}  \multicolumn{7}{c}{\textbf{\textit{Few-shot Prompting}}} \\
    Code LLM 62B & \quad \enspace \, \typedesc/ & 16.58 & 29.68 & 29.68 & 42.30 & 25.90 \\
    ~~$+$ \giftforcode/ & \quad \enspace \, \typedesc/ & 21.52 & 36.73 & 33.58 & 48.61 & 27.35 \\
    \midrule
    Code LLM 62B  & \quad \ioexample/ & 19.85 & 33.77 & 30.79 & 43.23 & 26.41 \\
    ~~$+$ \giftforcode/ & \quad \ioexample/& 25.23 & 42.30 & 38.03 & 53.99 & 28.66 \\
    \midrule
    Code LLM 62B & \quad \iosummary/ & 23.75 & 37.11 & 34.50 & 46.75 & 26.25 \\
    ~~$+$ \giftforcode/ & \quad \iosummary/ & \textbf{28.01} & \textbf{43.79} & \textbf{43.04} & \textbf{55.47} & \textbf{29.34} \\

    \bottomrule
    \end{tabular}
    \caption{\passat{k} on \arcade/ and \dsonek/.
    For each type of I/O specification in \cref{tab:output_spec:examples} (\eg~$+$\iosummary/), intents are  augmented with I/O specifications of that type (\eg~intents inline with I/O summary) in fine-tuning data or few-shot exemplars. At test time, input intents use the same type of I/O specifications.}
    \label{tab:exp:arcade_result}
\end{table*}

\subsection{Setup}
\label{subsec:experiment_setup}

\paragraph{Base Code LLM}
We use a \textsc{PaLM} based, decoder-only code language model with 62B parameters~\citep{Chowdhery2022PaLMSL}. The model was first pre-trained on a collection of 1.3T tokens of web documents and github code data, and was then fine-tuned on a disjoint set of 64B Python code tokens together with 10B tokens from Python Jupyter notebooks. The fine-tuning process, including the \giftforcode/ procedure, can be conducted through the fine-tuning API on Google cloud Vertex AI.
Notably, this base model has strong performance in data science programming tasks, outperforming the StarCoder-15B Python model on \arcade/, as evidenced in \cref{tab:exp:arcade_result}
\footnote{StarCoder 15B has slightly higher \passat{1} on \dsonek/ compared to numbers in~\citet{Ni2023L2CEvalEL}.}.

\paragraph{Learning Methods}
We evaluated the performance of both the baseline and instruction-tuned models across a range of data formats, as shown in~\cref{tab:exp:arcade_result}. For each I/O specification type, we augmented the intents and few-shot exemplars with specifications of the corresponding type. Similarly, at test time, we augmented the intents with the same type of I/O specifications. For the few-shot prompting, we manually created exemplars for all types of specifications. These exemplars were prepended to the prompt when querying the LLM for code generation.

\paragraph{Simulate Noisy I/O Specifications at Test Time}
At testing time, the generation of \iosummary/ underwent a minor modification from the process detailed in~\cref{subsec:code_execution_and_io_inference}. We remove the concrete input/output variable states $\{ \var \}$ to produce noisy I/O summaries, simulating scenarios where users might give noisy I/O specifications~\citep{Devlin2017RobustFillNP}. We illustrate an example in \cref{sec:appendix_iosummary_examples} where the LLM generates an imperfect specification. 
While the \enquote{generalist} LLM (\palm2/) uses the code solution to generate noisy I/O summaries, we remark that the code LLM, which we aim to evaluate, does not have access to the ground truth solution. In other words, the \enquote{generalist} LLM acts merely as a \enquote{data labeler} to create I/O summaries in prompts in order to construct the evaluation dataset.
It is also a common practice in program synthesis to derive specifications from ground truth solutions, which then serve as the sole input to the model during its evaluation~\citep{Balog2016DeepCoderLT}.

\paragraph{Metrics}
Developing unit tests for data science programming tasks is often non-trivial. For example, creating a test for a data wrangling task like \enquote{What are the most populous cities in each country} (as shown in \cref{fig:intro_teaser}) is not feasible.
In the \arcade/ dataset, each question is accompanied by an annotated program.
Then the evaluation uses a set of heuristics to approximate whether the execution output of a predicted program and the annotated reference are functionally equivalent, by comparing their outputs.
On the \dsonek/ dataset, evaluation relies on the provided tests. 

We adopted the \passat{k} metrics as defined in~\citet{Chen2021EvaluatingLL,austin2021lambdacode}, which is calculated as the fraction of problems with at least one correct sample given $k$ samples. Following~\citet{yin22acl}, we drew 50 samples to calculate \passat{5} and \passat{20} to reduce the variance in \arcade/.
Similar to~\citet{Lai2022DS1000}, we drew 40 samples to calculate \passat{1} on \dsonek/. Consistent with the original works' settings, the sampling temperature was set to 0.8 for \arcade/ and to 0.2 for \dsonek/ respectively.

\subsection{Main Results}
\label{subsec:main_results}

\begin{figure*}[t]
    \centering
    \includegraphics[width=0.9\columnwidth]{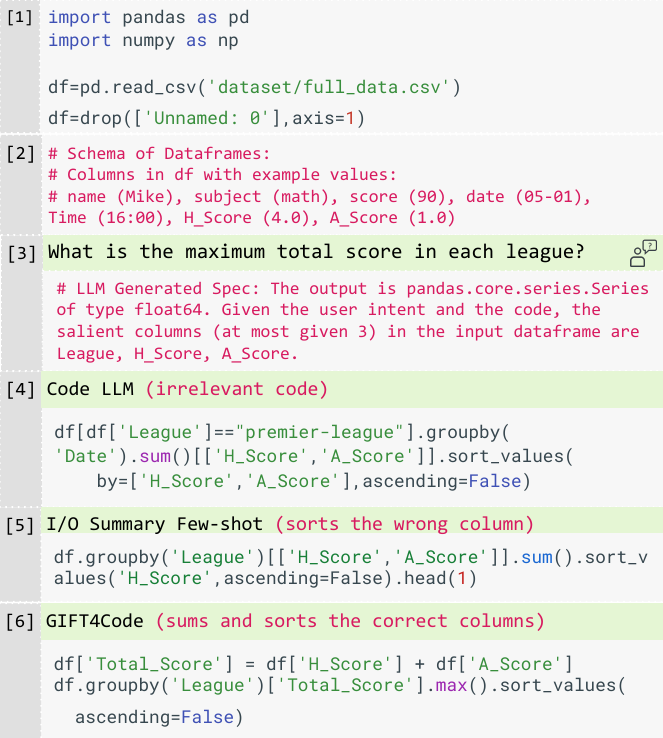}
    \includegraphics[width=0.9\columnwidth]{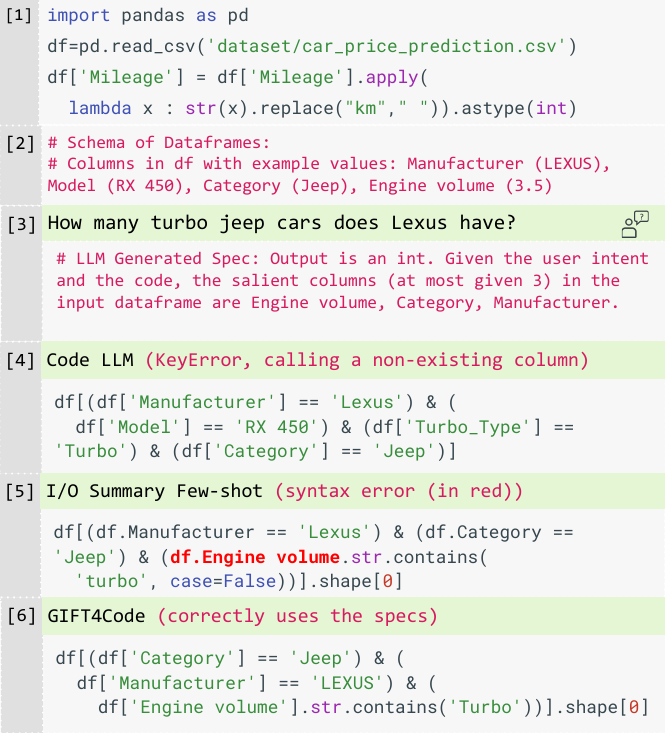}
    \caption{Two examples on \arcade/. \textbf{Left}: the base \nbmodel/ does not follow the intent to group the ``League'' column. Its few-shot prompting variant failed to correctly utilize the specifications. \giftforcode/'s solution aligns to the user's intent. \textbf{Right}: \nbmodel/ tries to call a non-exsiting column, leading to a \texttt{KeyError}. Its few-shot variant follows the specifications incorrectly, leading to the syntax error. \giftforcode/ generates correct solution. The prompt always includes dataframe schema, regardless of whether specifications are provided.}
    \label{fig:ex1ex2}
    \vspace{-0.5em}
\end{figure*}

\cref{tab:exp:arcade_result} presents the \passat{k} results on \arcade/ and \dsonek/. We evaluate both few-shot prompting and fine-tuning with synthetic data.
Specifically, for \arcade/ we evaluate on two versions of the dataset.
First, we consider the original version where an intent is prefixed by prior notebook cells as its programmatic context (\textbf{Full Context}), as well as a \textbf{No Context} ablation to simulate the scenario where users query a code LLM using an intent without any context. This no-context setting is more challenging, where the zero-shot performance of the base code LLM is nearly halved. The standard errors in all cells of the table are less than 0.5\%, and are excluded for clarity in presentation.

In our few-shot prompting experiments, we observe that \passat{k} generally improves with more detailed I/O specifications.
Interestingly, on \arcade/, the improvements from prompting using I/O specifications compared to the baseline where no I/O specifications were used (\textbf{no spec}), are more notable in the more challenging no-context scenario (\eg~$12.45 \mapsto 23.75$ \textit{v.s.}~$27.75 \mapsto 37.11$ for $+$\iosummary/). 
This trend suggests that additional specifications could provide more valuable clarifications when adequate programmatic contexts are lacking.

Next, we fine-tune the base code LLM using our synthetic parallel data with different types of I/O specifications. 
Even without using any I/O specifications in the synthetic data, \giftforcode/ already registers significant improvements on \arcade/ (\passat{20}: ~$37.47 \mapsto 46.94$).
The model-predicted code solutions are filtered using executability heuristics, which helps improve the quality of the synthetic data, and a model fine-tuned on such data could generally be better at following users' intents, even without I/O specifications.
Moreover, by fine-tuning the model to follow intents with additional I/O specifications, we observe significantly better results. With I/O summaries ($+$\iosummary/), \giftforcode/ improves \passat{20} from $46.75$ to $55.47$.
We also remark that instruction fine-tuning using natural language I/O summaries yields the best results on both datasets.
Intuitively, those I/O summaries could encode salient information in target input and output variables through natural language descriptions, which could make it easier for the model to capture patterns in the data.

\subsection{Qualitative Analysis}
\label{subsec:qualitative_analysis}
To gain deeper insights into the behavior of the different models, we present a qualitative analysis of the baseline model, its few-shot prompting variant with LLM specifications (\iosummary/), and our proposed \giftforcode/ approach. We illustrate this analysis using two examples from the \arcade/ dataset, as shown in \cref{fig:ex1ex2}.

In the first example (\cref{fig:ex1ex2}, \textit{Left}), the base \nbmodel/ (cell 4) fails to group the \enquote{League} column as requested in the user's intent. Note that the provided solution is still executable so it cannot be filtered by executability heuristics.
The few-shot prompting variant with I/O summary (cell 5) also fails here.
It struggles to correctly utilize these specifications despite selecting the correct salient columns, leading to an output that does not meet the user's requirement either.
In contrast, the output from \giftforcode/ (cell 6) successfully generates a solution which computes the sum of the two salient columns then sorts the result, effectively utilizing the provided specifications and adhering to the user's intent.

The second example (\cref{fig:ex1ex2}, \textit{Right}) further underscores the advantages of \giftforcode/.
The baseline \nbmodel/ (cell 4) attempts to call a non-existing column (\texttt{Turbo\_Type}), leading to a \texttt{KeyError}. This represents a failure mode that the model tries to adhere a user's intent but generates an inexecutable solution that refers to incorrect input variables due to lack of I/O specifications.
The few-shot prompting variant (cell 5) presents another interesting failure mode.
While the model is trying to follow the additional I/O specification (presumably because of the few-shot demonstrations) by referring to the \texttt{Engine volume} column in the specification, it fails to generate a syntactically correct code snippet (\textcolor{red}{\texttt{df.Engine volume}}).
It is important to note that this is a common failure mode of the few-shot prompting model, as we explain in~\cref{fig:exec_vs_error} later.
Once again, \giftforcode/ outperforms other settings, generating solutions that answer the natural language question while following specifications.

\subsection{Execution Rate vs \bigpassat{k}}
\label{subsec:executability_vs_pass@k}

We delve further into the experimental results to examine the relationship between executability and the quality of the generated code. Surprisingly, we observe that a model with a higher execution rate does not necessarily produce better code.
\cref{fig:exec_vs_error} plots the frequency of common error types  alongside the code execution rates of different models' predictions.
The baseline \nbmodel/, despite generating a substantial amount of executable code (higher $\bigstar$), often produces incorrect (irrelevant) solutions, leading to a high executable rate but low \passat{k} accuracies (\cref{tab:exp:arcade_result}).
This suggests that a model's ability to generate executable code does not necessarily indicate its competence in generating semantically correct code that aligns with the user's intent.
This insight is further evidenced when we fine-tune the model on synthetic data without I/O specifications, labeled as \textbf{SyntheticFT} in~\cref{fig:exec_vs_error}.
The model’s execution rate decreases in this scenario because it attempts to better align with user intent, leading to a higher rate of schema understanding errors (\eg~referring to non-existing columns as in cell 4, Example 2, \cref{fig:ex1ex2}).

\begin{figure}
  \begin{center}
    \includegraphics[width=0.45\textwidth]{figures/exec_vs_error.pdf}
  \end{center}
  \vspace{-10pt}
  \caption{Frequency of error types and code execution rate for different methods. Bottom x-axis stands for the counts of schema errors and syntax errors. Top x-axis represents execution rate. Instruction fine-tuning without specifications (SyntheticFT) decreases executability. Few-shot prompting with specifications reduces schema understanding errors with more syntax errors. \giftforcode/ achieves the best performance by combining their strengths.}
  \label{fig:exec_vs_error}
  \vspace{-0.5em}
\end{figure}
Incorporating I/O specifications using few-shot prompting leads to another interesting observation. We observe a reduction in schema understanding errors, indicating that the model indeed benefits from the extra specifications. However, given the vast diversity of Python I/O specifications, it is impossible to cover all variants within the few-shot exemplars. Consequently, the model struggles to reliably leverage the specifications, leading to a surge in syntax errors when referring to arbitrarily string-valued schema elements (\eg~cell 5, Example 2, \cref{fig:ex1ex2}).
\giftforcode/ effectively mitigates these syntax errors, leading to a higher execution rate while achieving the best \passat{k} accuracies.

\section{Related Work}
\label{sec:related_work}

\textbf{Execution Guided Code Generation}\quad 
One area of study primarily focuses on utilizing execution as I/O examples, facilitating the synthesis of programs that align with the intended behavior. \citet{Gulwani2016ProgrammingBE} involves synthesizing intended programs in an underlying domain-specific language (DSL) from example based specifications. This method has been further explored and adapted to different applications in subsequent studies~\citep{Devlin2017RobustFillNP,Chen2018ExecutionGuidedNP,Bunel2018LeveragingGA}. 
Another strand of research~\citep{Chen2021LatentEF,Wang2018RobustTG,Ellis2019WriteEA} leverages intermediate execution results to guide the search of programs.
More recently, there have been attempts to utilize program execution results to verify and select code samples predicted by LLMs, either during auto-regressive decoding to prune search space~\citep{Zhang2023PlanningWL}, or by few-shot prompting \citep{Chen2023TeachingLL} and post-hoc reranking~\citep{Shi2022NaturalLT,Ni2023LEVERLT}. CodeT~\citep{Chen2022CodeTCG} was proposed to generate tests then to conduct dual execution agreements to filter the generated solutions during the inference time.

\textbf{Instruction Fine-tuning}\quad
Instruction fine-tuning is a widely adopted approach to address the misalignment issue in LLM-generated content. LLMs such as FLAN~\citep{Wei2021FinetunedLM}, which excel at understanding and executing instructions from prompts, are trained on labeled training data.  Reinforcement learning with human feedback (RLHF) aims to mitigate the amount of labeling effort using model-based reward \citep{Ouyang2022TrainingLM}.
Other works also confirmed the effectiveness of using instructional data in the fine-tuning stage~\citep{Mishra2021CrossTaskGV,Sanh2021MultitaskPT,Chung2022ScalingIL,Wang2022SuperNaturalInstructionsGV}.
To lower labeling cost, several recent works explored the possibility of automatic instruction generation~\citep{Ye2022GuessTI, Zhou2022LargeLM, Honovich2022InstructionIF}.
\textsc{STAR}~\citep{Zelikman2022Parsel} bootstraps the model's reasoning ability using self-generated rationales. 
Our work differs from this line by considering execution-based specifications.
Additionally, recent works attempted to distill instruction following data from more capable LLMs that have already been instruction-tuned~\citep{Honovich2022UnnaturalIT,alpaca,vicuna2023,Peng2023InstructionTW}.
In contrast, \giftforcode/ generates synthetic data from vanilla LLMs that have not gone through instruction-tunning.

\textbf{Synthetic Data from LLMs}\quad
Besides generating data for instruction following, a number of recent studies have also harnessed general-purpose LLMs to generate realistic synthetic data in areas where labeled data limited, such as language understanding and clinical research \citep{Rosenbaum2022CLASPFC, Tang2023DoesSD, Borisov2022LanguageMA, Liu2022AugmentingMT, Rosenbaum2022LINGUISTLM, Josifoski2023ExploitingAF}.
To improve the quality of synthetic data extracted from LLMs, such approaches usually apply a rejection sampling procedure and filter predictions based on domain-specific heuristics such as logical consistency~\citep{Bhagavatula2022I2D2IK,yin22acl}. 
\giftforcode/ is in spirit of this line in that it leverages program execution feedback to filter code predictions~\citep{Xu2020AutoQAFD}.

\section{Conclusion}
\label{sec:conclusion}
We have presented \giftforcode/, a framework for instruction fine-tuning large language models of code in which the training is guided by execution based specifications. Empirically, we demonstrated how our approach enhances the quality of generated code by following user-provided specifications, substantially improving accuracy on two challenging data science benchmarks, \arcade/ and \dsonek/.

\bibliography{ref}

\begin{thebibliography}{57}
\providecommand{\natexlab}[1]{#1}
\providecommand{\url}[1]{\texttt{#1}}
\expandafter\ifx\csname urlstyle\endcsname\relax
  \providecommand{\doi}[1]{doi: #1}\else
  \providecommand{\doi}{doi: \begingroup \urlstyle{rm}\Url}\fi

\bibitem[Alur et~al.(2013)Alur, Bod{\'i}k, Juniwal, Martin, Raghothaman, Seshia, Singh, Solar-Lezama, Torlak, and Udupa]{Alur2013SyntaxguidedS}
Alur, R., Bod{\'i}k, R., Juniwal, G., Martin, M. M.~K., Raghothaman, M., Seshia, S.~A., Singh, R., Solar-Lezama, A., Torlak, E., and Udupa, A.
\newblock Syntax-guided synthesis.
\newblock \emph{2013 Formal Methods in Computer-Aided Design}, pp.\  1--8, 2013.

\bibitem[Austin et~al.(2021)Austin, Odena, Nye, Bosma, Michalewski, Dohan, Jiang, Cai, Terry, Le, et~al.]{austin2021lambdacode}
Austin, J., Odena, A., Nye, M., Bosma, M., Michalewski, H., Dohan, D., Jiang, E., Cai, C., Terry, M., Le, Q., et~al.
\newblock Program synthesis with large language models.
\newblock \emph{arXiv preprint arXiv:2108.07732}, 2021.

\bibitem[Balog et~al.(2016)Balog, Gaunt, Brockschmidt, Nowozin, and Tarlow]{Balog2016DeepCoderLT}
Balog, M., Gaunt, A.~L., Brockschmidt, M., Nowozin, S., and Tarlow, D.
\newblock Deepcoder: Learning to write programs.
\newblock \emph{ArXiv}, abs/1611.01989, 2016.

\bibitem[Berant et~al.(2013)Berant, Chou, Frostig, and Liang]{berant2013freebase}
Berant, J., Chou, A., Frostig, R., and Liang, P.
\newblock Semantic parsing on {F}reebase from question-answer pairs.
\newblock In \emph{Empirical Methods in Natural Language Processing (EMNLP)}, 2013.

\bibitem[Bhagavatula et~al.(2022)Bhagavatula, Hwang, Downey, Bras, Lu, Sakaguchi, Swayamdipta, West, and Choi]{Bhagavatula2022I2D2IK}
Bhagavatula, C., Hwang, J.~D., Downey, D., Bras, R.~L., Lu, X., Sakaguchi, K., Swayamdipta, S., West, P., and Choi, Y.
\newblock I2d2: Inductive knowledge distillation with neurologic and self-imitation.
\newblock \emph{ArXiv}, abs/2212.09246, 2022.

\bibitem[Bird et~al.(2023)Bird, Ford, Zimmermann, Forsgren, Kalliamvakou, Lowdermilk, and Gazit]{bird23flight}
Bird, C., Ford, D., Zimmermann, T., Forsgren, N., Kalliamvakou, E., Lowdermilk, T., and Gazit, I.
\newblock Taking flight with copilot: Early insights and opportunities of ai-powered pair-programming tools.
\newblock \emph{Queue}, 20\penalty0 (6):\penalty0 35–57, jan 2023.
\newblock ISSN 1542-7730.
\newblock \doi{10.1145/3582083}.
\newblock URL \url{https://doi.org/10.1145/3582083}.

\bibitem[Borisov et~al.(2022)Borisov, Sessler, Leemann, Pawelczyk, and Kasneci]{Borisov2022LanguageMA}
Borisov, V., Sessler, K., Leemann, T., Pawelczyk, M., and Kasneci, G.
\newblock Language models are realistic tabular data generators.
\newblock \emph{ArXiv}, abs/2210.06280, 2022.

\bibitem[Bunel et~al.(2018)Bunel, Hausknecht, Devlin, Singh, and Kohli]{Bunel2018LeveragingGA}
Bunel, R., Hausknecht, M.~J., Devlin, J., Singh, R., and Kohli, P.
\newblock Leveraging grammar and reinforcement learning for neural program synthesis.
\newblock \emph{ArXiv}, abs/1805.04276, 2018.

\bibitem[Chen et~al.(2022)Chen, Zhang, Nguyen, Zan, Lin, Lou, and Chen]{Chen2022CodeTCG}
Chen, B., Zhang, F., Nguyen, A., Zan, D., Lin, Z., Lou, J.-G., and Chen, W.
\newblock Codet: Code generation with generated tests.
\newblock \emph{ArXiv}, abs/2207.10397, 2022.
\newblock URL \url{https://api.semanticscholar.org/CorpusID:250920542}.

\bibitem[Chen et~al.(2021{\natexlab{a}})Chen, Tworek, Jun, Yuan, Ponde, Kaplan, Edwards, Burda, Joseph, Brockman, Ray, Puri, Krueger, Petrov, Khlaaf, Sastry, Mishkin, Chan, Gray, Ryder, Pavlov, Power, Kaiser, Bavarian, Winter, Tillet, Such, Cummings, Plappert, Chantzis, Barnes, Herbert-Voss, Guss, Nichol, Babuschkin, Balaji, Jain, Carr, Leike, Achiam, Misra, Morikawa, Radford, Knight, Brundage, Murati, Mayer, Welinder, McGrew, Amodei, McCandlish, Sutskever, and Zaremba]{Chen2021EvaluatingLL}
Chen, M., Tworek, J., Jun, H., Yuan, Q., Ponde, H., Kaplan, J., Edwards, H., Burda, Y., Joseph, N., Brockman, G., Ray, A., Puri, R., Krueger, G., Petrov, M., Khlaaf, H., Sastry, G., Mishkin, P., Chan, B., Gray, S., Ryder, N., Pavlov, M., Power, A., Kaiser, L., Bavarian, M., Winter, C., Tillet, P., Such, F.~P., Cummings, D.~W., Plappert, M., Chantzis, F., Barnes, E., Herbert-Voss, A., Guss, W.~H., Nichol, A., Babuschkin, I., Balaji, S.~A., Jain, S., Carr, A., Leike, J., Achiam, J., Misra, V., Morikawa, E., Radford, A., Knight, M.~M., Brundage, M., Murati, M., Mayer, K., Welinder, P., McGrew, B., Amodei, D., McCandlish, S., Sutskever, I., and Zaremba, W.
\newblock Evaluating large language models trained on code.
\newblock \emph{ArXiv}, abs/2107.03374, 2021{\natexlab{a}}.

\bibitem[Chen et~al.(2018)Chen, Liu, and Song]{Chen2018ExecutionGuidedNP}
Chen, X., Liu, C., and Song, D.~X.
\newblock Execution-guided neural program synthesis.
\newblock In \emph{International Conference on Learning Representations}, 2018.

\bibitem[Chen et~al.(2021{\natexlab{b}})Chen, Song, and Tian]{Chen2021LatentEF}
Chen, X., Song, D.~X., and Tian, Y.
\newblock Latent execution for neural program synthesis beyond domain-specific languages.
\newblock In \emph{Neural Information Processing Systems}, 2021{\natexlab{b}}.

\bibitem[Chen et~al.(2023)Chen, Lin, Sch{\"a}rli, and Zhou]{Chen2023TeachingLL}
Chen, X., Lin, M., Sch{\"a}rli, N., and Zhou, D.
\newblock Teaching large language models to self-debug.
\newblock \emph{ArXiv}, abs/2304.05128, 2023.

\bibitem[Chiang et~al.(2023)Chiang, Li, Lin, Sheng, Wu, Zhang, Zheng, Zhuang, Zhuang, Gonzalez, Stoica, and Xing]{vicuna2023}
Chiang, W.-L., Li, Z., Lin, Z., Sheng, Y., Wu, Z., Zhang, H., Zheng, L., Zhuang, S., Zhuang, Y., Gonzalez, J.~E., Stoica, I., and Xing, E.~P.
\newblock Vicuna: An open-source chatbot impressing gpt-4 with 90\%* chatgpt quality, March 2023.
\newblock URL \url{https://lmsys.org/blog/2023-03-30-vicuna/}.

\bibitem[Chowdhery et~al.(2022)Chowdhery, Narang, Devlin, Bosma, Mishra, Roberts, Barham, Chung, Sutton, Gehrmann, Schuh, Shi, Tsvyashchenko, Maynez, Rao, Barnes, Tay, Shazeer, Prabhakaran, Reif, Du, Hutchinson, Pope, Bradbury, Austin, Isard, Gur-Ari, Yin, Duke, Levskaya, Ghemawat, Dev, Michalewski, Garc{\'i}a, Misra, Robinson, Fedus, Zhou, Ippolito, Luan, Lim, Zoph, Spiridonov, Sepassi, Dohan, Agrawal, Omernick, Dai, Pillai, Pellat, Lewkowycz, Moreira, Child, Polozov, Lee, Zhou, Wang, Saeta, D{\'i}az, Firat, Catasta, Wei, Meier-Hellstern, Eck, Dean, Petrov, and Fiedel]{Chowdhery2022PaLMSL}
Chowdhery, A., Narang, S., Devlin, J., Bosma, M., Mishra, G., Roberts, A., Barham, P., Chung, H.~W., Sutton, C., Gehrmann, S., Schuh, P., Shi, K., Tsvyashchenko, S., Maynez, J., Rao, A., Barnes, P., Tay, Y., Shazeer, N.~M., Prabhakaran, V., Reif, E., Du, N., Hutchinson, B.~C., Pope, R., Bradbury, J., Austin, J., Isard, M., Gur-Ari, G., Yin, P., Duke, T., Levskaya, A., Ghemawat, S., Dev, S., Michalewski, H., Garc{\'i}a, X., Misra, V., Robinson, K., Fedus, L., Zhou, D., Ippolito, D., Luan, D., Lim, H., Zoph, B., Spiridonov, A., Sepassi, R., Dohan, D., Agrawal, S., Omernick, M., Dai, A.~M., Pillai, T.~S., Pellat, M., Lewkowycz, A., Moreira, E., Child, R., Polozov, O., Lee, K., Zhou, Z., Wang, X., Saeta, B., D{\'i}az, M., Firat, O., Catasta, M., Wei, J., Meier-Hellstern, K.~S., Eck, D., Dean, J., Petrov, S., and Fiedel, N.
\newblock Palm: Scaling language modeling with pathways.
\newblock \emph{ArXiv}, abs/2204.02311, 2022.

\bibitem[Chung et~al.(2022)Chung, Hou, Longpre, Zoph, Tay, Fedus, Li, Wang, Dehghani, Brahma, Webson, Gu, Dai, Suzgun, Chen, Chowdhery, Valter, Narang, Mishra, Yu, Zhao, Huang, Dai, Yu, Petrov, hsin Chi, Dean, Devlin, Roberts, Zhou, Le, and Wei]{Chung2022ScalingIL}
Chung, H.~W., Hou, L., Longpre, S., Zoph, B., Tay, Y., Fedus, W., Li, E., Wang, X., Dehghani, M., Brahma, S., Webson, A., Gu, S.~S., Dai, Z., Suzgun, M., Chen, X., Chowdhery, A., Valter, D., Narang, S., Mishra, G., Yu, A.~W., Zhao, V., Huang, Y., Dai, A.~M., Yu, H., Petrov, S., hsin Chi, E.~H., Dean, J., Devlin, J., Roberts, A., Zhou, D., Le, Q.~V., and Wei, J.
\newblock Scaling instruction-finetuned language models.
\newblock \emph{ArXiv}, abs/2210.11416, 2022.

\bibitem[Devlin et~al.(2017)Devlin, Uesato, Bhupatiraju, Singh, rahman Mohamed, and Kohli]{Devlin2017RobustFillNP}
Devlin, J., Uesato, J., Bhupatiraju, S., Singh, R., rahman Mohamed, A., and Kohli, P.
\newblock Robustfill: Neural program learning under noisy i/o.
\newblock \emph{ArXiv}, abs/1703.07469, 2017.

\bibitem[Ellis et~al.(2019)Ellis, Nye, Pu, Sosa, Tenenbaum, and Solar-Lezama]{Ellis2019WriteEA}
Ellis, K., Nye, M., Pu, Y., Sosa, F., Tenenbaum, J.~B., and Solar-Lezama, A.
\newblock Write, execute, assess: Program synthesis with a repl.
\newblock In \emph{Neural Information Processing Systems}, 2019.

\bibitem[Fried et~al.(2022)Fried, Aghajanyan, Lin, Wang, Wallace, Shi, Zhong, Yih, Zettlemoyer, and Lewis]{fried2022incoder}
Fried, D., Aghajanyan, A., Lin, J., Wang, S., Wallace, E., Shi, F., Zhong, R., Yih, W.-t., Zettlemoyer, L., and Lewis, M.
\newblock Incoder: A generative model for code infilling and synthesis.
\newblock \emph{arXiv preprint arXiv:2204.05999}, 2022.

\bibitem[Google(2023)]{palm2}
Google.
\newblock {PaLM2} technical report.
\newblock \url{https://ai.google/static/documents/palm2techreport.pdf}, 2023.
\newblock Accessed: 2023-05-10.

\bibitem[Gulwani(2011)]{Gulwani2011AutomatingSP}
Gulwani, S.
\newblock Automating string processing in spreadsheets using input-output examples.
\newblock In \emph{ACM-SIGACT Symposium on Principles of Programming Languages}, 2011.
\newblock URL \url{https://api.semanticscholar.org/CorpusID:886323}.

\bibitem[Gulwani(2016)]{Gulwani2016ProgrammingBE}
Gulwani, S.
\newblock Programming by examples - and its applications in data wrangling.
\newblock In \emph{Dependable Software Systems Engineering}, 2016.

\bibitem[Gulwani et~al.(2012)Gulwani, Harris, and Singh]{Gulwani2012SpreadsheetDM}
Gulwani, S., Harris, W.~R., and Singh, R.
\newblock Spreadsheet data manipulation using examples.
\newblock \emph{Commun. ACM}, 55:\penalty0 97--105, 2012.

\bibitem[Gulwani et~al.(2015)Gulwani, Hern{\'a}ndez-Orallo, Kitzelmann, Muggleton, Schmid, and Zorn]{gulwani2015inductive}
Gulwani, S., Hern{\'a}ndez-Orallo, J., Kitzelmann, E., Muggleton, S.~H., Schmid, U., and Zorn, B.
\newblock Inductive programming meets the real world.
\newblock \emph{Communications of the ACM}, 58\penalty0 (11):\penalty0 90--99, 2015.

\bibitem[Honovich et~al.(2022{\natexlab{a}})Honovich, Scialom, Levy, and Schick]{Honovich2022UnnaturalIT}
Honovich, O., Scialom, T., Levy, O., and Schick, T.
\newblock Unnatural instructions: Tuning language models with (almost) no human labor.
\newblock \emph{ArXiv}, abs/2212.09689, 2022{\natexlab{a}}.

\bibitem[Honovich et~al.(2022{\natexlab{b}})Honovich, Shaham, Bowman, and Levy]{Honovich2022InstructionIF}
Honovich, O., Shaham, U., Bowman, S.~R., and Levy, O.
\newblock Instruction induction: From few examples to natural language task descriptions.
\newblock \emph{ArXiv}, abs/2205.10782, 2022{\natexlab{b}}.

\bibitem[Iyer et~al.(2018)Iyer, Konstas, Cheung, and Zettlemoyer]{Iyer2018MappingLT}
Iyer, S., Konstas, I., Cheung, A., and Zettlemoyer, L.
\newblock Mapping language to code in programmatic context.
\newblock \emph{ArXiv}, abs/1808.09588, 2018.

\bibitem[Jain et~al.(2022)Jain, Vaidyanath, Iyer, Natarajan, Parthasarathy, Rajamani, and Sharma]{jain2022jigsaw}
Jain, N., Vaidyanath, S., Iyer, A., Natarajan, N., Parthasarathy, S., Rajamani, S., and Sharma, R.
\newblock Jigsaw: Large language models meet program synthesis.
\newblock In \emph{Proceedings of the 44th International Conference on Software Engineering}, pp.\  1219--1231, 2022.

\bibitem[Josifoski et~al.(2023)Josifoski, Sakota, Peyrard, and West]{Josifoski2023ExploitingAF}
Josifoski, M., Sakota, M., Peyrard, M., and West, R.
\newblock Exploiting asymmetry for synthetic training data generation: Synthie and the case of information extraction.
\newblock \emph{ArXiv}, abs/2303.04132, 2023.

\bibitem[Lai et~al.(2022)Lai, Li, Wang, Zhang, Zhong, Zettlemoyer, tau Yih, Fried, Wang, and Yu]{Lai2022DS1000}
Lai, Y., Li, C., Wang, Y., Zhang, T., Zhong, R., Zettlemoyer, L., tau Yih, S.~W., Fried, D., Wang, S., and Yu, T.
\newblock Ds-1000: A natural and reliable benchmark for data science code generation.
\newblock \emph{ArXiv}, abs/2211.11501, 2022.

\bibitem[Li et~al.(2023)Li, Allal, Zi, Muennighoff, Kocetkov, Mou, Marone, Akiki, Li, Chim, Liu, Zheltonozhskii, Zhuo, Wang, Dehaene, Davaadorj, Lamy-Poirier, Monteiro, Shliazhko, Gontier, Meade, Zebaze, Yee, Umapathi, Zhu, Lipkin, Oblokulov, Wang, Murthy, Stillerman, Patel, Abulkhanov, Zocca, Dey, Zhang, Fahmy, Bhattacharyya, Yu, Singh, Luccioni, Villegas, Kunakov, Zhdanov, Romero, Lee, Timor, Ding, Schlesinger, Schoelkopf, Ebert, Dao, Mishra, Gu, Robinson, Anderson, Dolan-Gavitt, Contractor, Reddy, Fried, Bahdanau, Jernite, Ferrandis, Hughes, Wolf, Guha, von Werra, and de~Vries]{Li2023StarCoderMT}
Li, R., Allal, L.~B., Zi, Y., Muennighoff, N., Kocetkov, D., Mou, C., Marone, M., Akiki, C., Li, J., Chim, J., Liu, Q., Zheltonozhskii, E., Zhuo, T.~Y., Wang, T., Dehaene, O., Davaadorj, M., Lamy-Poirier, J., Monteiro, J., Shliazhko, O., Gontier, N., Meade, N., Zebaze, A., Yee, M.-H., Umapathi, L.~K., Zhu, J., Lipkin, B., Oblokulov, M., Wang, Z., Murthy, R., Stillerman, J., Patel, S.~S., Abulkhanov, D., Zocca, M., Dey, M., Zhang, Z., Fahmy, N., Bhattacharyya, U., Yu, W., Singh, S., Luccioni, S., Villegas, P., Kunakov, M., Zhdanov, F., Romero, M., Lee, T., Timor, N., Ding, J., Schlesinger, C., Schoelkopf, H., Ebert, J., Dao, T., Mishra, M., Gu, A., Robinson, J., Anderson, C.~J., Dolan-Gavitt, B., Contractor, D., Reddy, S., Fried, D., Bahdanau, D., Jernite, Y., Ferrandis, C.~M., Hughes, S.~M., Wolf, T., Guha, A., von Werra, L., and de~Vries, H.
\newblock Starcoder: may the source be with you!
\newblock 2023.

\bibitem[Li et~al.(2022)Li, Choi, Chung, Kushman, Schrittwieser, Leblond, Tom, Eccles, Keeling, Gimeno, Lago, Hubert, Choy, de, d’Autume, Babuschkin, Chen, Huang, Welbl, Gowal, Alexey, Cherepanov, Molloy, Mankowitz, Robson, Kohli, de, Freitas, Kavukcuoglu, and Vinyals]{Li2022CompetitionlevelCG}
Li, Y., Choi, D.~H., Chung, J., Kushman, N., Schrittwieser, J., Leblond, R., Tom, Eccles, Keeling, J., Gimeno, F., Lago, A.~D., Hubert, T., Choy, P., de, C., d’Autume, M., Babuschkin, I., Chen, X., Huang, P.-S., Welbl, J., Gowal, S., Alexey, Cherepanov, Molloy, J., Mankowitz, D.~J., Robson, E.~S., Kohli, P., de, N., Freitas, Kavukcuoglu, K., and Vinyals, O.
\newblock Competition-level code generation with alphacode.
\newblock \emph{Science}, 378:\penalty0 1092 -- 1097, 2022.

\bibitem[Liu et~al.(2022)Liu, Ye, Yu, Blunsom, and Song]{Liu2022AugmentingMT}
Liu, Q., Ye, Z., Yu, T., Blunsom, P., and Song, L.
\newblock Augmenting multi-turn text-to-sql datasets with self-play.
\newblock In \emph{Conference on Empirical Methods in Natural Language Processing}, 2022.

\bibitem[Mishra et~al.(2021)Mishra, Khashabi, Baral, and Hajishirzi]{Mishra2021CrossTaskGV}
Mishra, S., Khashabi, D., Baral, C., and Hajishirzi, H.
\newblock Cross-task generalization via natural language crowdsourcing instructions.
\newblock In \emph{Annual Meeting of the Association for Computational Linguistics}, 2021.

\bibitem[Ni et~al.(2023{\natexlab{a}})Ni, Iyer, Radev, Stoyanov, tau Yih, Wang, and Lin]{Ni2023LEVERLT}
Ni, A., Iyer, S., Radev, D.~R., Stoyanov, V., tau Yih, W., Wang, S.~I., and Lin, X.~V.
\newblock Lever: Learning to verify language-to-code generation with execution.
\newblock \emph{ArXiv}, abs/2302.08468, 2023{\natexlab{a}}.

\bibitem[Ni et~al.(2023{\natexlab{b}})Ni, Yin, Zhao, Riddell, Feng, Shen, Yin, Liu, Yavuz, Xiong, Joty, Zhou, Radev, and Cohan]{Ni2023L2CEvalEL}
Ni, A., Yin, P., Zhao, Y., Riddell, M., Feng, T., Shen, R., Yin, S., Liu, Y., Yavuz, S., Xiong, C., Joty, S.~R., Zhou, Y., Radev, D.~R., and Cohan, A.
\newblock L2ceval: Evaluating language-to-code generation capabilities of large language models.
\newblock \emph{ArXiv}, abs/2309.17446, 2023{\natexlab{b}}.
\newblock URL \url{https://api.semanticscholar.org/CorpusID:263310373}.

\bibitem[Nijkamp et~al.(2022)Nijkamp, Pang, Hayashi, Tu, Wang, Zhou, Savarese, and Xiong]{Nijkamp2022CodeGenAO}
Nijkamp, E., Pang, B., Hayashi, H., Tu, L., Wang, H., Zhou, Y., Savarese, S., and Xiong, C.
\newblock Codegen: An open large language model for code with multi-turn program synthesis.
\newblock 2022.

\bibitem[Ouyang et~al.(2022)Ouyang, Wu, Jiang, Almeida, Wainwright, Mishkin, Zhang, Agarwal, Slama, Ray, Schulman, Hilton, Kelton, Miller, Simens, Askell, Welinder, Christiano, Leike, and Lowe]{Ouyang2022TrainingLM}
Ouyang, L., Wu, J., Jiang, X., Almeida, D., Wainwright, C.~L., Mishkin, P., Zhang, C., Agarwal, S., Slama, K., Ray, A., Schulman, J., Hilton, J., Kelton, F., Miller, L.~E., Simens, M., Askell, A., Welinder, P., Christiano, P.~F., Leike, J., and Lowe, R.~J.
\newblock Training language models to follow instructions with human feedback.
\newblock \emph{ArXiv}, abs/2203.02155, 2022.

\bibitem[Peng et~al.(2023)Peng, Li, He, Galley, and Gao]{Peng2023InstructionTW}
Peng, B., Li, C., He, P., Galley, M., and Gao, J.
\newblock Instruction tuning with gpt-4.
\newblock \emph{ArXiv}, abs/2304.03277, 2023.

\bibitem[Rosenbaum et~al.(2022{\natexlab{a}})Rosenbaum, Soltan, Hamza, Saffari, Damonte, and Groves]{Rosenbaum2022CLASPFC}
Rosenbaum, A., Soltan, S., Hamza, W., Saffari, A., Damonte, M., and Groves, I.
\newblock Clasp: Few-shot cross-lingual data augmentation for semantic parsing.
\newblock In \emph{AACL}, 2022{\natexlab{a}}.

\bibitem[Rosenbaum et~al.(2022{\natexlab{b}})Rosenbaum, Soltan, Hamza, Versley, and Boese]{Rosenbaum2022LINGUISTLM}
Rosenbaum, A., Soltan, S., Hamza, W., Versley, Y., and Boese, M.
\newblock Linguist: Language model instruction tuning to generate annotated utterances for intent classification and slot tagging.
\newblock In \emph{International Conference on Computational Linguistics}, 2022{\natexlab{b}}.

\bibitem[Rozi{\`e}re et~al.(2023)Rozi{\`e}re, Gehring, Gloeckle, Sootla, Gat, Tan, Adi, Liu, Remez, Rapin, Kozhevnikov, Evtimov, Bitton, Bhatt, Ferrer, Grattafiori, Xiong, D'efossez, Copet, Azhar, Touvron, Martin, Usunier, Scialom, and Synnaeve]{Rozire2023CodeLO}
Rozi{\`e}re, B., Gehring, J., Gloeckle, F., Sootla, S., Gat, I., Tan, X., Adi, Y., Liu, J., Remez, T., Rapin, J., Kozhevnikov, A., Evtimov, I., Bitton, J., Bhatt, M.~P., Ferrer, C.~C., Grattafiori, A., Xiong, W., D'efossez, A., Copet, J., Azhar, F., Touvron, H., Martin, L., Usunier, N., Scialom, T., and Synnaeve, G.
\newblock Code llama: Open foundation models for code.
\newblock \emph{ArXiv}, abs/2308.12950, 2023.
\newblock URL \url{https://api.semanticscholar.org/CorpusID:261100919}.

\bibitem[Sanh et~al.(2021)Sanh, Webson, Raffel, Bach, Sutawika, Alyafeai, Chaffin, Stiegler, Scao, Raja, Dey, Bari, Xu, Thakker, Sharma, Szczechla, Kim, Chhablani, Nayak, Datta, Chang, Jiang, Wang, Manica, Shen, Yong, Pandey, Bawden, Wang, Neeraj, Rozen, Sharma, Santilli, F{\'e}vry, Fries, Teehan, Biderman, Gao, Bers, Wolf, and Rush]{Sanh2021MultitaskPT}
Sanh, V., Webson, A., Raffel, C., Bach, S.~H., Sutawika, L., Alyafeai, Z., Chaffin, A., Stiegler, A., Scao, T.~L., Raja, A., Dey, M., Bari, M.~S., Xu, C., Thakker, U., Sharma, S., Szczechla, E., Kim, T., Chhablani, G., Nayak, N.~V., Datta, D., Chang, J., Jiang, M. T.-J., Wang, H., Manica, M., Shen, S., Yong, Z.~X., Pandey, H., Bawden, R., Wang, T., Neeraj, T., Rozen, J., Sharma, A., Santilli, A., F{\'e}vry, T., Fries, J.~A., Teehan, R., Biderman, S.~R., Gao, L., Bers, T., Wolf, T., and Rush, A.~M.
\newblock Multitask prompted training enables zero-shot task generalization.
\newblock \emph{ArXiv}, abs/2110.08207, 2021.

\bibitem[Shi et~al.(2022)Shi, Fried, Ghazvininejad, Zettlemoyer, and Wang]{Shi2022NaturalLT}
Shi, F., Fried, D., Ghazvininejad, M., Zettlemoyer, L., and Wang, S.~I.
\newblock Natural language to code translation with execution.
\newblock \emph{ArXiv}, abs/2204.11454, 2022.

\bibitem[Shi et~al.(2020)Shi, Bieber, and Singh]{Shi2020TFCoderPS}
Shi, K., Bieber, D., and Singh, R.
\newblock Tf-coder: Program synthesis for tensor manipulations.
\newblock \emph{ACM Transactions on Programming Languages and Systems (TOPLAS)}, 44:\penalty0 1 -- 36, 2020.
\newblock URL \url{https://api.semanticscholar.org/CorpusID:214605958}.

\bibitem[Tang et~al.(2023)Tang, Han, Jiang, and Hu]{Tang2023DoesSD}
Tang, R., Han, X., Jiang, X., and Hu, X.
\newblock Does synthetic data generation of llms help clinical text mining?
\newblock \emph{ArXiv}, abs/2303.04360, 2023.

\bibitem[Taori et~al.(2023)Taori, Gulrajani, Zhang, Dubois, Li, Guestrin, Liang, and Hashimoto]{alpaca}
Taori, R., Gulrajani, I., Zhang, T., Dubois, Y., Li, X., Guestrin, C., Liang, P., and Hashimoto, T.~B.
\newblock Stanford alpaca: An instruction-following llama model.
\newblock \url{https://github.com/tatsu-lab/stanford_alpaca}, 2023.

\bibitem[Wang et~al.(2018)Wang, Tatwawadi, Brockschmidt, Huang, Mao, Polozov, and Singh]{Wang2018RobustTG}
Wang, C., Tatwawadi, K., Brockschmidt, M., Huang, P.-S., Mao, Y., Polozov, O., and Singh, R.
\newblock Robust text-to-sql generation with execution-guided decoding.
\newblock \emph{arXiv: Computation and Language}, 2018.

\bibitem[Wang et~al.(2022{\natexlab{a}})Wang, Kordi, Mishra, Liu, Smith, Khashabi, and Hajishirzi]{Wang2022SelfInstructAL}
Wang, Y., Kordi, Y., Mishra, S., Liu, A., Smith, N.~A., Khashabi, D., and Hajishirzi, H.
\newblock Self-instruct: Aligning language model with self generated instructions.
\newblock \emph{ArXiv}, abs/2212.10560, 2022{\natexlab{a}}.

\bibitem[Wang et~al.(2022{\natexlab{b}})Wang, Mishra, Alipoormolabashi, Kordi, Mirzaei, Arunkumar, Ashok, Dhanasekaran, Naik, Stap, Pathak, Karamanolakis, Lai, Purohit, Mondal, Anderson, Kuznia, Doshi, Patel, Pal, Moradshahi, Parmar, Purohit, Varshney, Kaza, Verma, Puri, Karia, Sampat, Doshi, Mishra, Reddy, Patro, Dixit, Shen, Baral, Choi, Smith, Hajishirzi, and Khashabi]{Wang2022SuperNaturalInstructionsGV}
Wang, Y., Mishra, S., Alipoormolabashi, P., Kordi, Y., Mirzaei, A., Arunkumar, A., Ashok, A., Dhanasekaran, A.~S., Naik, A., Stap, D., Pathak, E., Karamanolakis, G., Lai, H.~G., Purohit, I., Mondal, I., Anderson, J., Kuznia, K., Doshi, K., Patel, M., Pal, K.~K., Moradshahi, M., Parmar, M., Purohit, M., Varshney, N., Kaza, P.~R., Verma, P., Puri, R.~S., Karia, R., Sampat, S.~K., Doshi, S., Mishra, S.~D., Reddy, S., Patro, S., Dixit, T., Shen, X., Baral, C., Choi, Y., Smith, N.~A., Hajishirzi, H., and Khashabi, D.
\newblock Super-naturalinstructions: Generalization via declarative instructions on 1600+ nlp tasks.
\newblock In \emph{Conference on Empirical Methods in Natural Language Processing}, 2022{\natexlab{b}}.

\bibitem[Wei et~al.(2021)Wei, Bosma, Zhao, Guu, Yu, Lester, Du, Dai, and Le]{Wei2021FinetunedLM}
Wei, J., Bosma, M., Zhao, V., Guu, K., Yu, A.~W., Lester, B., Du, N., Dai, A.~M., and Le, Q.~V.
\newblock Finetuned language models are zero-shot learners.
\newblock \emph{ArXiv}, abs/2109.01652, 2021.

\bibitem[Xu et~al.(2020)Xu, Semnani, Campagna, and Lam]{Xu2020AutoQAFD}
Xu, S., Semnani, S.~J., Campagna, G., and Lam, M.~S.
\newblock Autoqa: From databases to q\&a semantic parsers with only synthetic training data.
\newblock \emph{ArXiv}, abs/2010.04806, 2020.

\bibitem[Ye et~al.(2022)Ye, Kim, Jang, Shin, and Seo]{Ye2022GuessTI}
Ye, S., Kim, D., Jang, J., Shin, J., and Seo, M.
\newblock Guess the instruction! flipped learning makes language models stronger zero-shot learners.
\newblock \emph{ArXiv}, abs/2210.02969, 2022.

\bibitem[Yin et~al.(2022)Yin, Wieting, Sil, and Neubig]{yin22acl}
Yin, P., Wieting, J.~F., Sil, A., and Neubig, G.
\newblock On the ingredients of an effective zero-shot semantic parser.
\newblock In \emph{Annual Conference of the Association for Computational Linguistics (ACL)}, Dublin, Ireland, May 2022.
\newblock URL \url{https://arxiv.org/abs/2110.08381}.

\bibitem[Zelikman et~al.(2022)Zelikman, Huang, Poesia, Goodman, and Haber]{Zelikman2022Parsel}
Zelikman, E., Huang, Q., Poesia, G., Goodman, N.~D., and Haber, N.
\newblock Parsel: A (de-)compositional framework for algorithmic reasoning with language models.
\newblock 2022.

\bibitem[Zhang et~al.(2023)Zhang, Chen, Shen, Ding, Tenenbaum, and Gan]{Zhang2023PlanningWL}
Zhang, S., Chen, Z., Shen, Y., Ding, M., Tenenbaum, J.~B., and Gan, C.
\newblock Planning with large language models for code generation.
\newblock \emph{ArXiv}, abs/2303.05510, 2023.

\bibitem[Zhou et~al.(2022)Zhou, Muresanu, Han, Paster, Pitis, Chan, and Ba]{Zhou2022LargeLM}
Zhou, Y., Muresanu, A.~I., Han, Z., Paster, K., Pitis, S., Chan, H., and Ba, J.
\newblock Large language models are human-level prompt engineers.
\newblock \emph{ArXiv}, abs/2211.01910, 2022.

\end{thebibliography}
\bibliographystyle{ref}

\newpage
\appendix
\onecolumn

\section{Applying \giftforcode/ to Data Science Code Generation}
\label{sec:appendix_gift4code}

In this appendix section, we demonstrate the practical application of our proposed approach \giftforcode/, as discussed in \cref{sec:GIFT4Code}, to a specific dataset, \arcade/~\citep{yin22acl}. We follow the same setup as described in \cref{sec:GIFT4Code}. Our starting point involves a \enquote{Generalist LLM} (\palm2/) and a Code LLM (PaLM based 62B model in our case), the alignment of which we seek to improve.

\subsection{Synthetic Data Generation}
\label{subsec:appendix:synthetic_data_generation}

We first gathered CSV files from diverse GitHub repositories focusing on data science projects. These CSV files encompass a wide range of data types, structures, and domains, serving as the programmatic context to initialize the synthetic data generation. Subsequently, we employed a \enquote{Generalist} \palm2/ LLM specifically designed for natural language understanding tasks, distinguished from the base code LLMs. This model was utilized to generate natural language questions based on the information extracted from the collected CSV files. For the following sections, we denote the set of natural number from 0 to $W$ and $\N_{W}=\{ 1, 2, 3, \dots, W\}$. 

\paragraph{Creating NL Intents (Questions)}
Using the entire CSV file as the programmatic context to prompt the \enquote{Generalist} \palm2/ LLM is infeasible due to its length after tokenization. Instead, we extracted the header for each CSV file as the programmatic context $\context_i$ to query the LLM to generate the NL intents $\{ \intent_i \}$. 
Given the length of an entire CSV file after tokenization, using it as the programmatic context to prompt the \enquote{generalist} LLM is impractical. Instead, we derive the header and its first three rows from each CSV file as the programmatic context, denoted as $\context_i$, to query the LLM to produce NL intents ${\intent_i}$.
To construct the few-shot exemplars, we randomly selected a subset of CSV headers, denoted as $\contextset'=\{ \context_i' \mid i\in \N_P \}$ where $P < 10$.
The prime symbol $'$ in the superscript denotes the variable will be used in the few-shot exemplars. We manually crafted a set of 10 data science-related questions corresponding to each $\context_i'$, denoted as $\intentset_i'=\{ \intent_i^j \mid j\in \N_{10} \}$.
This process allowed us to create a few-shot prompt set $\promptset$ that consists of pairs of CSV headers and associated questions, formulated as $\promptset = \{ (\context_i', \intentset_i')\}_{i=1}^P$.

We employed the standard few-shot prompting approach, which concatenates $\promptset$ to each CSV header $\context_i$ when querying the LLM. In specific, each header $\context_i$ is augmented to $\hat{\context_i} = (\promptset, \context_i)$.
In this configuration, the LLM is prompted to generate questions related to $\context_i$, while emulating the style of \promptset.
After generation, we perform diversity filtering on $\intentset_i$. Let $\widehat{\intentset_i}$ denote the set of NL questions after the filtering process. We initialize $\widehat{\intentset_i}$ as $\{q_0 \}$ where $q_0\sim \mathcal{U}(\intentset_i)$), a question randomly selected from $\intentset_i$. Similar to~\citet{Wang2022SelfInstructAL}, we iterate over $\{\intentset_i \setminus q_0 \}$, a new question is added to the set $\widehat{\intentset_i}$ only when its ROUGE-L overlap with any $q\in\widehat{\intentset_i}$ is less than 0.7. 
Questions that are repetitive are also filtered out. For notation simplicity, in the following sections, we use $\intentset_i$ to represent the set of NL intents after the filtering process. Empirically, we observe that the NL questions generated by this LLM encompassed a wide range of tasks relevant to the given programmatic context, as shown in \cref{fig:appendix:generated_NL_intent_ex1} and \cref{fig:appendix:generated_NL_intent_ex2}.

\paragraph{Predicting Code Solutions}
Subsequently, we employed the code LLM, which we aim to improve, to generate code that addresses these NL questions. For each question $q_i^j\in \intentset_i$ where $j$ denotes the $j^{\text{th}}$ question associated with the programmatic context $\context_i$, we can draw code solutions from $P_{\textrm{LLM}}( \code \mid \context_i, q_i^j)$. This leads to a pool of code samples $\{ \code\}_i^j$.
Following this, we applied heuristic filtering on these code samples, adopting two criteria: 1) Maximizing the diversity of Python library API calls, and 2) Ensuring the produced execution output is meaningful, for instance, yielding a \texttt{pandas.DataFrame} object if the code is executable. 
This process resulted in a synthetic dataset $\SyntheticDataset$ that enables instruction fine-tuning. Notice that $\SyntheticDataset$ does not contain any I/O specifications yet.

The synthetic data generation process leverages the capabilities of the LLM to create a diverse set of natural language intents that correspond to the header information extracted from the CSV files. By incorporating the knowledge and understanding of \enquote{Generalist} \palm2/ LLM, we are able to generate a rich and varied synthetic dataset that contains a wide range of data science tasks and scenarios.

\paragraph{Executing Code Samples}
Executing code derived from a pre-training corpus which was used to train the code LLM can be challenging, as it often demands notoriously complicated environment setup. In contrast, the synthetic code samples offer a significant advantage, as they can be executed more easily without the need for complex environment setup. Moreover, the execution of synthetic code samples enables us to track variables and their states both before and after running the code. This information can be utilized to identify input variables that exhibit state changes after execution. We label these variables as inputs to the code. In addition, by examining the final states of these input variables, we can effectively establish the input-output specifications of the code.

\section{LLM Generated NL Intents}
\label{sec:appendix_llm_generated_nl_intents}

In this section, we demonstrated the NL questions generated by the \enquote{generalist} \palm2/ LLM on the \arcade/ dataset. To begin, we provide an exemplar used in the few-shot prompt when querying the LLM to generate NL questions, as seen in \cref{fig:append:fewshot_prompt_to_generate_nl_intents}. The given exemplar consists of the CSV header along with the first three rows. If any entry within the first three rows exceeds 50 characters in its string representation, we truncate it to maintain conciseness. As shown in \cref{fig:append:fewshot_prompt_to_generate_nl_intents}, we handcrafted 10 diverse questions, covering as many data science topics as possible.

In this section, we provide two examples of NL intent generation. Each of these examples, as shown in \cref{fig:appendix:generated_NL_intent_ex1} and \cref{fig:appendix:generated_NL_intent_ex2}, includes both the programmatic context and the output generated by the LLM. \cref{fig:appendix:generated_NL_intent_ex1} illustrates an instance regarding a Pokémon game experience. Notably, the LLM tends to generate relatively basic questions in this context, which don't necessitate the use of complex Python APIs such as \texttt{pandas.groupby}. Conversely, \cref{fig:appendix:generated_NL_intent_ex2} presents an example related to a Netflix TV and movie dataset. In this case, the LLM produces more comprehensive questions. Addressing these questions requires multiple API calls, indicating a higher level of complexity.

\begin{figure}
\begin{Verbatim}[commandchars=\\\{\}]
First 3 rows from dataset 4\PYGZhy{}wheeler\PYGZhy{}EV.csv (column data types in parentheses)
| Comfort (float) | Performance (float) | Fuel Economy (float)
| Value for Money (float) | Condition (string) | driven (string) | rating (float)
| model\PYGZus{}name (string) |
|\PYGZhy{}\PYGZhy{}\PYGZhy{}\PYGZhy{}\PYGZhy{}\PYGZhy{}\PYGZhy{}\PYGZhy{}\PYGZhy{}\PYGZhy{}\PYGZhy{}\PYGZhy{}\PYGZhy{}\PYGZhy{}\PYGZhy{}\PYGZhy{}\PYGZhy{}\PYGZhy{}\PYGZhy{}\PYGZhy{}\PYGZhy{}\PYGZhy{}\PYGZhy{}\PYGZhy{}\PYGZhy{}\PYGZhy{}\PYGZhy{}\PYGZhy{}\PYGZhy{}\PYGZhy{}\PYGZhy{}\PYGZhy{}\PYGZhy{}\PYGZhy{}\PYGZhy{}\PYGZhy{}\PYGZhy{}\PYGZhy{}\PYGZhy{}\PYGZhy{}\PYGZhy{}\PYGZhy{}\PYGZhy{}\PYGZhy{}\PYGZhy{}\PYGZhy{}\PYGZhy{}\PYGZhy{}\PYGZhy{}\PYGZhy{}\PYGZhy{}\PYGZhy{}\PYGZhy{}\PYGZhy{}\PYGZhy{}\PYGZhy{}\PYGZhy{}\PYGZhy{}\PYGZhy{}\PYGZhy{}\PYGZhy{}\PYGZhy{}\PYGZhy{}\PYGZhy{}\PYGZhy{}\PYGZhy{}\PYGZhy{}\PYGZhy{}\PYGZhy{}\PYGZhy{}\PYGZhy{}\PYGZhy{}\PYGZhy{}\PYGZhy{}\PYGZhy{}\PYGZhy{}\PYGZhy{}\PYGZhy{}\PYGZhy{}\PYGZhy{}|
| 4 | 5 | 5 | 5 | New | Few hundred kilometers | 5 | hyundai kona |
| 1 | 1 | 1 | 1 | New | Haven\PYGZsq{}t driven it | 0 | hyundai kona |
| 5 | 5 | 5 | 4 | New | Few thousand kilometers | 5 | hyundai kona |

First 3 rows from dataset 2\PYGZhy{}wheeler\PYGZhy{}EV.csv (column data types in parentheses)
| Owned for (string) | Ridden for (string) | rating (int)
| Visual Appeal (float) | Reliability (float) | Performance (float)
| Service Experience (float) | Extra Features (float) | Comfort (float)
| Maintenance cost (float) | Value for Money (float) | Model Name (string) |
|\PYGZhy{}\PYGZhy{}\PYGZhy{}\PYGZhy{}\PYGZhy{}\PYGZhy{}\PYGZhy{}\PYGZhy{}\PYGZhy{}\PYGZhy{}\PYGZhy{}\PYGZhy{}\PYGZhy{}\PYGZhy{}\PYGZhy{}\PYGZhy{}\PYGZhy{}\PYGZhy{}\PYGZhy{}\PYGZhy{}\PYGZhy{}\PYGZhy{}\PYGZhy{}\PYGZhy{}\PYGZhy{}\PYGZhy{}\PYGZhy{}\PYGZhy{}\PYGZhy{}\PYGZhy{}\PYGZhy{}\PYGZhy{}\PYGZhy{}\PYGZhy{}\PYGZhy{}\PYGZhy{}\PYGZhy{}\PYGZhy{}\PYGZhy{}\PYGZhy{}\PYGZhy{}\PYGZhy{}\PYGZhy{}\PYGZhy{}\PYGZhy{}\PYGZhy{}\PYGZhy{}\PYGZhy{}\PYGZhy{}\PYGZhy{}\PYGZhy{}\PYGZhy{}\PYGZhy{}\PYGZhy{}\PYGZhy{}\PYGZhy{}\PYGZhy{}\PYGZhy{}\PYGZhy{}\PYGZhy{}\PYGZhy{}\PYGZhy{}\PYGZhy{}\PYGZhy{}\PYGZhy{}\PYGZhy{}\PYGZhy{}\PYGZhy{}\PYGZhy{}\PYGZhy{}\PYGZhy{}\PYGZhy{}\PYGZhy{}\PYGZhy{}\PYGZhy{}\PYGZhy{}\PYGZhy{}\PYGZhy{}\PYGZhy{}\PYGZhy{}|
| Never owned | nan | 1 | 3 | 4 | nan | nan | nan | 4 | nan | 1 | TVS iQube |
| \PYGZgt{} 1 yr | \PYGZlt{} 5000 kms | 1 | 3 | 1 | nan | 1 | nan | 3 | nan | 3 | TVS iQube |
| \PYGZlt{} 3 months | \PYGZlt{} 5000 kms | 3 | 4 | 4 | nan | 2 | nan | 5 | nan | 2 | TVS iQube |

Here are a series of contextually dependent data wrangling and exploratory data
analysis tasks for the dataset:
Task 1: How many new cars have a performance rating of more than 3?
Task 2: Among the Hyundai Kona reviews, do those with a purchase history have
average performance rating better than the ones that have not yet been purchased?
Task 3: Create a new column called \PYGZdq{}Review\PYGZsq{}s Word Count\PYGZdq{} and Show the rating of
the shortest review
Task 4: Which model has the highest average word count?
Task 5: How many reviews are for used cars for this model?
Task 6: What are the top five models with most number of bikes having mileage
less than 5000 kilometers
Task 7: Which of these models has the highest comfort score on average?
Task 8: What are those models mostly used for?
Task 9: How many times the word scooter was mentioned in the bike\PYGZsq{}s reviews?
Task 10: Show a list of words that were mentioned only once in the reviews
[END]
\end{Verbatim}
\captionof{listing}{An exemplar in the few-shot prompt when querying LLM to generate NL intents.}
\label{fig:append:fewshot_prompt_to_generate_nl_intents}
\end{figure}

\begin{figure}
\begin{subfigure}[t!]{1.0\textwidth}
\centering
\begin{Verbatim}[commandchars=\\\{\}]
| id (int) | identifier (string) | species\PYGZus{}id (int) | height (int) | weight (int)
| base\PYGZus{}experience (int) | order (int) | is\PYGZus{}default (int) |
|\PYGZhy{}\PYGZhy{}\PYGZhy{}\PYGZhy{}\PYGZhy{}\PYGZhy{}\PYGZhy{}\PYGZhy{}\PYGZhy{}\PYGZhy{}\PYGZhy{}\PYGZhy{}\PYGZhy{}\PYGZhy{}\PYGZhy{}\PYGZhy{}\PYGZhy{}\PYGZhy{}\PYGZhy{}\PYGZhy{}\PYGZhy{}\PYGZhy{}\PYGZhy{}\PYGZhy{}\PYGZhy{}\PYGZhy{}\PYGZhy{}\PYGZhy{}\PYGZhy{}\PYGZhy{}\PYGZhy{}\PYGZhy{}\PYGZhy{}\PYGZhy{}\PYGZhy{}\PYGZhy{}\PYGZhy{}\PYGZhy{}\PYGZhy{}\PYGZhy{}\PYGZhy{}\PYGZhy{}\PYGZhy{}\PYGZhy{}\PYGZhy{}\PYGZhy{}\PYGZhy{}\PYGZhy{}\PYGZhy{}\PYGZhy{}\PYGZhy{}\PYGZhy{}\PYGZhy{}\PYGZhy{}\PYGZhy{}\PYGZhy{}\PYGZhy{}\PYGZhy{}\PYGZhy{}\PYGZhy{}\PYGZhy{}\PYGZhy{}\PYGZhy{}\PYGZhy{}\PYGZhy{}\PYGZhy{}\PYGZhy{}\PYGZhy{}\PYGZhy{}\PYGZhy{}\PYGZhy{}\PYGZhy{}\PYGZhy{}\PYGZhy{}\PYGZhy{}|
| 1 | bulbasaur | 1 | 7 | 69 | 64 | 1 | 1 |
| 2 | ivysaur | 2 | 10 | 130 | 142 | 2 | 1 |
| 3 | venusaur | 3 | 20 | 1000 | 263 | 3 | 1 |
\end{Verbatim}
\caption{The programmatic context to query the \enquote{generalist} \palm2/ LLM, albeit only a subset of the columns are displayed.}
\end{subfigure}

\begin{subfigure}[t!]{1.0\textwidth}
\centering
\begin{Verbatim}[commandchars=\\\{\}]
Here are a series of contextually dependent data wrangling and exploratory data
analysis tasks for the dataset:
Task 1: How many pokemons are there in the Pokedex?
Task 2: Find the total number of columns that are integers.
Task 3: Calculate the ratio of mean weight to height for each pokemon
Task 4: What is the weight of “Snorlax” ?
Task 5: How many pokemon have the same average base experience as their id?
Task 6: Find the order for each pokemon that weighs less than 100 pounds
Task 7: What is the “mean” and “median” of “height” column ?
Task 8: Show the names of the pokemons with minimum and maximum weight, height
and base experience.
Task 9: Show the first 20 and last 10 pokemon with their average base experience.
Task 10: Create a new column called \PYGZdq{}size\PYGZus{}cat\PYGZdq{} that has size categories for
pokemon (child: 1\PYGZhy{}10, teenager: 11\PYGZhy{}20, adult: 21+)
\end{Verbatim}
\caption{LLM generated questions for the above programmatic context.}
\end{subfigure}
\captionof{listing}{Programmatic context and generated questions on an example concerning pokemon game experience dataset.}
\label{fig:appendix:generated_NL_intent_ex1}
\end{figure}

\begin{figure}
\begin{subfigure}[t!]{1.0\textwidth}
\centering
\begin{Verbatim}[commandchars=\\\{\}]
| show\PYGZus{}id (string) | type (string) | title (string) | director (string) |
| country (string) | date\PYGZus{}added (string) | release\PYGZus{}year (int) | ... |
|\PYGZhy{}\PYGZhy{}\PYGZhy{}\PYGZhy{}\PYGZhy{}\PYGZhy{}\PYGZhy{}\PYGZhy{}\PYGZhy{}\PYGZhy{}\PYGZhy{}\PYGZhy{}\PYGZhy{}\PYGZhy{}\PYGZhy{}\PYGZhy{}\PYGZhy{}\PYGZhy{}\PYGZhy{}\PYGZhy{}\PYGZhy{}\PYGZhy{}\PYGZhy{}\PYGZhy{}\PYGZhy{}\PYGZhy{}\PYGZhy{}\PYGZhy{}\PYGZhy{}\PYGZhy{}\PYGZhy{}\PYGZhy{}\PYGZhy{}\PYGZhy{}\PYGZhy{}\PYGZhy{}\PYGZhy{}\PYGZhy{}\PYGZhy{}\PYGZhy{}\PYGZhy{}\PYGZhy{}\PYGZhy{}\PYGZhy{}\PYGZhy{}\PYGZhy{}\PYGZhy{}\PYGZhy{}\PYGZhy{}\PYGZhy{}\PYGZhy{}\PYGZhy{}\PYGZhy{}\PYGZhy{}\PYGZhy{}\PYGZhy{}\PYGZhy{}\PYGZhy{}\PYGZhy{}\PYGZhy{}\PYGZhy{}\PYGZhy{}\PYGZhy{}\PYGZhy{}\PYGZhy{}\PYGZhy{}\PYGZhy{}\PYGZhy{}\PYGZhy{}\PYGZhy{}\PYGZhy{}\PYGZhy{}\PYGZhy{}\PYGZhy{}\PYGZhy{}\PYGZhy{}\PYGZhy{}\PYGZhy{}\PYGZhy{}\PYGZhy{}|
| s1 | Movie | Dick Johnson Is Dead | Kirsten Johnson | United States |
    September 25, 2021 | 2020 | PG\PYGZhy{}13 | 90 min | ... |
| s2 | TV Show | Blood \PYGZam{} Water | nan | South Africa |
    September 24, 2021 | 2021 | TV\PYGZhy{}MA | 2 Seasons | ... |
| s3 | TV Show | Ganglands | Julien Leclercq | nan |
    September 24, 2021 | 2021 | TV\PYGZhy{}MA | 1 Season | ... |
\end{Verbatim}
\caption{The programmatic context to query the \enquote{generalist} \palm2/ LLM, albeit only a subset of the columns are displayed.}
\end{subfigure}

\begin{subfigure}[t!]{1.0\textwidth}
\centering
\begin{Verbatim}[commandchars=\\\{\}]
Here are a series of contextually dependent data wrangling and exploratory data
analysis tasks for the dataset:
Task 1: What is the count of number of seasons for each show?
Task 2: How Many shows have “TV14” as rating?
Task 3: How many movies have a rating of 13+?
Task 4: Show the top 10 TV shows with most number of seasons
Task 5: Make a new column \PYGZdq{}genre\PYGZdq{} that combines all genres into one column.
Task 6: Show the names of casts who have been in at least 5 shows?
Task 7: How many TV Shows are there that have been released before 2017?
Task 8: For each director, how many shows have been added in 2020?
Task 9: Show the movies that have director\PYGZsq{}s name with the letter \PYGZsq{}b\PYGZsq{} in it.
Task 10: Show the number of shows released before 2020 in the genre “Documentaries”?
Task 11: For each type, what are the average durations?
(Show dataframe that has type and duration as columns)
\end{Verbatim}
\caption{LLM generated questions for the above programmatic context.}
\end{subfigure}
\captionof{listing}{Programmatic context and generated questions on an example concerning Netflix TV show and movie dataset.}
\label{fig:appendix:generated_NL_intent_ex2}
\end{figure}

\section{I/O Summary Examples}
\label{sec:appendix_iosummary_examples}

To begin with, we showed an example of a few-shot exemplar used to prompt the \enquote{generalist} \palm2/ LLM in generating an I/O summary for the \arcade/ dataset, as detailed in section \ref{subsec:code_execution_and_io_inference}.
The exemplar in \cref{fig:appendix:fewshot_exemplar_iospec_generation} comprises an input dataframe schema, a Python code solution, execution output, and user intent.
The anticipated response to this prompt is an I/O summary, outlining the input-output variable names and their types. In this example, there is only one variable — ""alc"" which is a \texttt{pandas.DataFrame}. Next, the LLM is expected to give a succinct description on the salient input columns, as well as a brief summary of the example output columns. 

We present two examples of LLM generated I/O summaries on the \arcade/ dataset, as illustrated in \cref{fig:appendix:IOsummary_ex1} and \cref{fig:appendix:IOsummary_ex2}.
As mentioned in \cref{subsec:experiment_setup}, we deliberately obscure the execution output details when prompting the LLM to generate an I/O summary. This step helps to more realistically simulate user provided specifications.
Each example consists of its prompt we used to query the LLM for the I/O summary, the resulting example augmented by the generated I/O summary and the ground truth execution output which was never exposed to LLM.

The first example \cref{fig:appendix:IOsummary_ex1} focuses on a dataset detailing mobile phone information, with the user intent being to determine the quantity of different smartphones released each decade. The subsequent I/O summary generated by the LLM identifies the "smartphone" and "year" columns as the most relevant from the input dataframe and describes the output values as being of int64 type, with an index year. The generated I/O summary correctly describes the ground truth presented in the bottom of \cref{fig:appendix:IOsummary_ex1}.

In the second example \cref{fig:appendix:IOsummary_ex2}, the LLM is dealing with movie data where the user's intent is to chronologically list the worldwide gross of films released each month from November 2021 to June 2022. 
The corresponding I/O summary generated by the LLM correctly identified the salient columns in the input dataframe, ``Worldwide'' and ``Released Date''.
However, an observed discrepancy between the ground truth output and the I/O summary indicates that the LLM's generation is not entirely accurate for this specific example.

These two test examples on \arcade/ illustrate the LLM's capabilities in generating I/O summaries, while highlighting the potential discrepancies that may occur.

\section{Additional Experimental Results}
\label{sec:appendix_additional_results}

In this section, we offer two additional experiments that supplement the results presented in \cref{subsec:main_results}. We have demonstrated the \giftforcode/ model consistently outperforms in all types of I/O specifications. A natural follow-up question might be whether the instruction fine-tuning degrades the model's programming ability if no specifications are provided by the user.  To address this concern, we have conducted further experiments where an instruction tuned model is evaluated on the \arcade/ dataset, in the absence of any specification.

The results in \cref{tab:appendix:exp:arcade_result} demonstrate that while the instruction tuned model does perform slightly worse than the model fine-tuned on data with no specifications, the difference is marginal. 
In the \passat{5} and \passat{20} settings, both with no context and full context, the model's performance (all types of specifications) only decreases by most 2\% when compared with the model without specifications. This is expected as the discrepancy between the fine-tuning data and the testing data could lead to a minor regression in performance.
This marginal decrease in performance is counterbalanced by the significant improvement we previously observed in \cref{subsec:main_results}. 
Therefore, \giftforcode/ with I/O summary still remains the superior method for instruction fine-tuning.

\begin{figure}
\begin{Verbatim}[commandchars=\\\{\}]
\PYG{l+s+sd}{\PYGZdq{}\PYGZdq{}\PYGZdq{}}
\PYG{l+s+sd}{The input dataframe schema is:}
\PYG{l+s+sd}{Schema of Dataframes:}
\PYG{l+s+sd}{Columns in alc with example values:}
\PYG{l+s+sd}{country (Afghanistan), beer\PYGZus{}servings (0), spirit\PYGZus{}servings (0), wine\PYGZus{}servings (0),}
\PYG{l+s+sd}{total\PYGZus{}litres\PYGZus{}of\PYGZus{}pure\PYGZus{}alcohol (0.0), continent (AS)}
\PYG{l+s+sd}{\PYGZdq{}\PYGZdq{}\PYGZdq{}}

\PYG{c+c1}{\PYGZsh{} The Python solution is:}
\PYG{k+kn}{import} \PYG{n+nn}{pandas} \PYG{k}{as} \PYG{n+nn}{pd}
\PYG{n}{alc} \PYG{o}{=} \PYG{n}{pd}\PYG{o}{.}\PYG{n}{read\PYGZus{}csv}\PYG{p}{(}\PYG{l+s+s2}{\PYGZdq{}drinks.csv\PYGZdq{}}\PYG{p}{)}
\PYG{n}{alc}\PYG{p}{[}\PYG{l+s+s1}{\PYGZsq{}continent\PYGZsq{}}\PYG{p}{]}\PYG{o}{.}\PYG{n}{unique}\PYG{p}{()}
\PYG{n}{alc}\PYG{o}{.}\PYG{n}{groupby}\PYG{p}{(}\PYG{l+s+s1}{\PYGZsq{}continent\PYGZsq{}}\PYG{p}{)}\PYG{o}{.}\PYG{n}{agg}\PYG{p}{(\PYGZob{}}\PYG{l+s+s1}{\PYGZsq{}beer\PYGZus{}servings\PYGZsq{}}\PYG{p}{:} \PYG{n}{np}\PYG{o}{.}\PYG{n}{mean}\PYG{p}{\PYGZcb{})}\PYG{o}{.}\PYG{n}{sort\PYGZus{}values}\PYG{p}{(}
    \PYG{l+s+s1}{\PYGZsq{}beer\PYGZus{}servings\PYGZsq{}}\PYG{p}{,} \PYG{n}{ascending}\PYG{o}{=}\PYG{k+kc}{True}\PYG{p}{)}
\PYG{n}{alc}\PYG{o}{.}\PYG{n}{dropna}\PYG{p}{()}

\PYG{c+c1}{\PYGZsh{} The execution output is:}
\PYG{n}{alc}\PYG{p}{:}
\PYG{o}{|}  \PYG{n}{beer\PYGZus{}servings} \PYG{p}{(}\PYG{n+nb}{float}\PYG{p}{)} \PYG{o}{|}
\PYG{o}{|\PYGZhy{}\PYGZhy{}\PYGZhy{}\PYGZhy{}\PYGZhy{}\PYGZhy{}\PYGZhy{}\PYGZhy{}\PYGZhy{}\PYGZhy{}\PYGZhy{}\PYGZhy{}\PYGZhy{}\PYGZhy{}\PYGZhy{}\PYGZhy{}\PYGZhy{}\PYGZhy{}\PYGZhy{}\PYGZhy{}\PYGZhy{}\PYGZhy{}\PYGZhy{}\PYGZhy{}|}
\PYG{o}{|}   \PYG{l+m+mi}{102}   \PYG{o}{|}
\PYG{o}{|}    \PYG{l+m+mi}{20}   \PYG{o}{|}

\PYG{c+c1}{\PYGZsh{} The user intent is:}
\PYG{n}{Rank} \PYG{n}{the} \PYG{n}{continents} \PYG{n}{that} \PYG{n}{consume} \PYG{n}{the} \PYG{n}{least} \PYG{n}{amount} \PYG{n}{of} \PYG{n}{beer} \PYG{n}{on} \PYG{n}{average}\PYG{o}{.}

\PYG{c+c1}{\PYGZsh{} The I/O specification is:}
\PYG{n}{alc}\PYG{p}{:} \PYG{n}{a} \PYG{n}{pandas}\PYG{o}{.}\PYG{n}{core}\PYG{o}{.}\PYG{n}{DataFrame}\PYG{o}{.} \PYG{n}{Given} \PYG{n}{the} \PYG{n}{user} \PYG{n}{intent} \PYG{o+ow}{and} \PYG{n}{the} \PYG{n}{code}\PYG{p}{,} \PYG{n}{the} \PYG{n}{salient}
\PYG{n}{columns} \PYG{p}{(}\PYG{n}{at} \PYG{n}{most} \PYG{n}{given} \PYG{l+m+mi}{3}\PYG{p}{)} \PYG{o+ow}{in} \PYG{n}{the} \PYG{n+nb}{input} \PYG{n}{dataframe} \PYG{n}{are} \PYG{n}{beer\PYGZus{}servings}\PYG{p}{,} \PYG{n}{continent}\PYG{o}{.}
\PYG{n}{The} \PYG{n}{output} \PYG{n}{dataframe} \PYG{n}{has} \PYG{n}{columns} \PYG{p}{(}\PYG{n}{at} \PYG{n}{most} \PYG{n}{given} \PYG{l+m+mi}{3}\PYG{p}{)} \PYG{n}{such} \PYG{k}{as} \PYG{n}{beer\PYGZus{}servings}\PYG{o}{.}
\end{Verbatim}
\captionof{listing}{An exemplar regarding \texttt{pandas.DataFrame}.}
\label{fig:appendix:fewshot_exemplar_iospec_generation}
\end{figure}

\begin{figure}
\begin{subfigure}[t!]{1.0\textwidth}
\centering
\begin{Verbatim}[commandchars=\\\{\}]
\PYG{l+s+sd}{\PYGZdq{}\PYGZdq{}\PYGZdq{}}
\PYG{l+s+sd}{The input schema is:}
\PYG{l+s+sd}{\PYGZsh{} Schema of Dataframes:}
\PYG{l+s+sd}{\PYGZsh{} Columns in phones with example values:}
\PYG{l+s+sd}{\PYGZsh{} manufacturer (Nokia), model (1100), form (Bar), smartphone (No), year (2003)}
\PYG{l+s+sd}{\PYGZdq{}\PYGZdq{}\PYGZdq{}}

\PYG{c+c1}{\PYGZsh{} The Python solution is:}
\PYG{n}{yearly\PYGZus{}smartphones} \PYG{o}{=} \PYG{n}{phones}\PYG{o}{.}\PYG{n}{groupby}\PYG{p}{([}\PYG{l+s+s1}{\PYGZsq{}year\PYGZsq{}}\PYG{p}{,} \PYG{l+s+s1}{\PYGZsq{}smartphone\PYGZsq{}}\PYG{p}{],}
    \PYG{n}{as\PYGZus{}index}\PYG{o}{=}\PYG{k+kc}{False}\PYG{p}{)}\PYG{o}{.}\PYG{n}{size}\PYG{p}{()}\PYG{o}{.}\PYG{n}{pivot\PYGZus{}table}\PYG{p}{(}
        \PYG{n}{index}\PYG{o}{=}\PYG{l+s+s1}{\PYGZsq{}year\PYGZsq{}}\PYG{p}{,}\PYG{n}{columns}\PYG{o}{=}\PYG{l+s+s1}{\PYGZsq{}smartphone\PYGZsq{}}\PYG{p}{,} \PYG{n}{values}\PYG{o}{=}\PYG{l+s+s1}{\PYGZsq{}size\PYGZsq{}}\PYG{p}{)}\PYG{o}{.}\PYG{n}{fillna}\PYG{p}{(}\PYG{l+m+mi}{0}\PYG{p}{)}
\PYG{n}{yearly\PYGZus{}smartphones}\PYG{o}{.}\PYG{n}{groupby}\PYG{p}{((}\PYG{n}{yearly\PYGZus{}smartphones}\PYG{o}{.}\PYG{n}{index}\PYG{o}{//}\PYG{l+m+mi}{10}\PYG{p}{)}\PYG{o}{*}\PYG{l+m+mi}{10}\PYG{p}{)}\PYG{o}{.}\PYG{n}{Yes}\PYG{o}{.}\PYG{n}{sum}\PYG{p}{()}

\PYG{c+c1}{\PYGZsh{} The execution output is:}
\PYG{n}{\PYGZus{}\PYGZus{}output\PYGZus{}\PYGZus{}}\PYG{p}{:}
\PYG{n}{pandas}\PYG{o}{.}\PYG{n}{core}\PYG{o}{.}\PYG{n}{series}\PYG{o}{.}\PYG{n}{Series}

\PYG{c+c1}{\PYGZsh{} The user intent is:}
\PYG{n}{How} \PYG{n}{many} \PYG{n}{different} \PYG{n}{smartphones} \PYG{n}{were} \PYG{n}{released} \PYG{n}{each} \PYG{n}{decade}\PYG{err}{?}

\PYG{n}{The} \PYG{n}{I}\PYG{o}{/}\PYG{n}{O} \PYG{n}{specification} \PYG{o+ow}{is}\PYG{p}{:}
\end{Verbatim}
\caption{The prompt used to query LLM for I/O summary.}
\end{subfigure}

\begin{subfigure}[t!]{1.0\textwidth}
\centering
\begin{Verbatim}[commandchars=\\\{\}]
\PYG{c+c1}{\PYGZsh{} In[ ]:}
\PYG{k+kn}{import} \PYG{n+nn}{pandas} \PYG{k}{as} \PYG{n+nn}{pd}
\PYG{k+kn}{import} \PYG{n+nn}{numpy} \PYG{k}{as} \PYG{n+nn}{np}
\PYG{n}{phones} \PYG{o}{=} \PYG{n}{pd}\PYG{o}{.}\PYG{n}{read\PYGZus{}csv}\PYG{p}{(}\PYG{l+s+s1}{\PYGZsq{}dataset/best\PYGZhy{}selling\PYGZhy{}mobile\PYGZhy{}phones.csv\PYGZsq{}}\PYG{p}{)}
\PYG{c+c1}{\PYGZsh{} In[ ]:}
\PYG{c+c1}{\PYGZsh{} Schema of Dataframes:}
\PYG{c+c1}{\PYGZsh{} Columns in phones with example values:}
\PYG{c+c1}{\PYGZsh{} manufacturer (Nokia), model (1100), form (Bar), smartphone (No), year (2003)}
\PYG{c+c1}{\PYGZsh{} In[ ]:}
\PYG{n}{phones}\PYG{p}{[}\PYG{n}{phones}\PYG{o}{.}\PYG{n}{form}\PYG{o}{.}\PYG{n}{str}\PYG{o}{.}\PYG{n}{lower}\PYG{p}{()}\PYG{o}{.}\PYG{n}{str}\PYG{o}{.}\PYG{n}{contains}\PYG{p}{(}\PYG{l+s+s1}{\PYGZsq{}touchscreen\PYGZsq{}}\PYG{p}{)]}\PYG{o}{.}\PYG{n}{groupby}\PYG{p}{(}
    \PYG{l+s+s1}{\PYGZsq{}manufacturer\PYGZsq{}}\PYG{p}{)}\PYG{o}{.}\PYG{n}{model}\PYG{o}{.}\PYG{n}{nunique}\PYG{p}{()}\PYG{o}{.}\PYG{n}{idxmax}\PYG{p}{()}
\PYG{c+c1}{\PYGZsh{} In[ ]:}
\PYG{n}{year\PYGZus{}phones} \PYG{o}{=} \PYG{n}{phones}\PYG{p}{[}\PYG{n}{phones}\PYG{p}{[}\PYG{l+s+s1}{\PYGZsq{}year\PYGZsq{}}\PYG{p}{]} \PYG{o}{\PYGZgt{}=} \PYG{n}{phones}\PYG{p}{[}\PYG{l+s+s1}{\PYGZsq{}year\PYGZsq{}}\PYG{p}{]}\PYG{o}{.}\PYG{n}{max}\PYG{p}{()}\PYG{o}{\PYGZhy{}}\PYG{l+m+mi}{15}\PYG{p}{]}
\PYG{n}{year\PYGZus{}phones}\PYG{o}{.}\PYG{n}{groupby}\PYG{p}{([}\PYG{l+s+s1}{\PYGZsq{}year\PYGZsq{}}\PYG{p}{,}\PYG{l+s+s1}{\PYGZsq{}manufacturer\PYGZsq{}}\PYG{p}{,}\PYG{l+s+s1}{\PYGZsq{}form\PYGZsq{}}\PYG{p}{],} \PYG{n}{as\PYGZus{}index}\PYG{o}{=}\PYG{k+kc}{False}\PYG{p}{)}\PYG{o}{.}\PYG{n}{size}\PYG{p}{()}\PYG{o}{.}\PYG{n}{pivot}\PYG{p}{(}
    \PYG{n}{index}\PYG{o}{=}\PYG{p}{[}\PYG{l+s+s1}{\PYGZsq{}year\PYGZsq{}}\PYG{p}{,}\PYG{l+s+s1}{\PYGZsq{}manufacturer\PYGZsq{}}\PYG{p}{],} \PYG{n}{columns}\PYG{o}{=}\PYG{l+s+s1}{\PYGZsq{}form\PYGZsq{}}\PYG{p}{)}\PYG{o}{.}\PYG{n}{fillna}\PYG{p}{(}\PYG{l+m+mi}{0}\PYG{p}{)}

\PYG{c+c1}{\PYGZsh{} In[ ]:}
\PYG{n}{How} \PYG{n}{many} \PYG{n}{different} \PYG{n}{smartphones} \PYG{n}{were} \PYG{n}{released} \PYG{n}{each} \PYG{n}{decade}\PYG{err}{?}
\PYG{l+s+sd}{\PYGZdq{}\PYGZdq{}\PYGZdq{}}
\PYG{l+s+sd}{Input\PYGZhy{}output Summary:}
\PYG{l+s+sd}{\PYGZus{}\PYGZus{}output\PYGZus{}\PYGZus{}: a pandas.core.series.Series. Given the user intent and the code, the}
\PYG{l+s+sd}{salient columns (at most given 3) in the input dataframe are smartphone, year.}
\PYG{l+s+sd}{The output values are of type int64, with an index year. Here is my code solution:}
\PYG{l+s+sd}{\PYGZdq{}\PYGZdq{}\PYGZdq{}}
\PYG{c+c1}{\PYGZsh{} In[ ]:}
\end{Verbatim}
\caption{A test example on \arcade/ augmented with a LLM generated I/O summary.}
\end{subfigure}

\begin{subfigure}[t!]{1.0\textwidth}
\centering
\begin{Verbatim}[commandchars=\\\{\}]
year
1990     0.0
2000    12.0
2010    58.0
Name: Yes, dtype: float64
\end{Verbatim}
\caption{The ground truth of the output in the above test example.}
\label{subfig:appenidx:ioex1_groundtruth}
\end{subfigure}
\captionof{listing}{An example of an I/O summary generated by the LLM on \arcade/ for a \texttt{pandas.Series}.}
\label{fig:appendix:IOsummary_ex1}
\end{figure}

\begin{figure}
\begin{subfigure}[t!]{1.0\textwidth}
\centering
\begin{Verbatim}[commandchars=\\\{\}]
\PYG{l+s+sd}{\PYGZdq{}\PYGZdq{}\PYGZdq{}}
\PYG{l+s+sd}{The input schema is:}
\PYG{l+s+sd}{Schema of Dataframes:}
\PYG{l+s+sd}{Columns in df with example values:}
\PYG{l+s+sd}{Movie (JugJugg Jeeyo), Worldwide (74.5), India Net (50.24), India Gross (54.5),}
\PYG{l+s+sd}{Budget (100), Verdict (None), Movie Type (Bollywood), Released Date (24\PYGZhy{}Jun\PYGZhy{}22)}
\PYG{l+s+sd}{\PYGZdq{}\PYGZdq{}\PYGZdq{}}

\PYG{c+c1}{\PYGZsh{} The Python solution is:}
\PYG{n}{df\PYGZus{}t}\PYG{p}{[[}\PYG{l+s+s1}{\PYGZsq{}month\PYGZsq{}}\PYG{p}{,} \PYG{l+s+s1}{\PYGZsq{}Worldwide\PYGZsq{}}\PYG{p}{]]}\PYG{o}{.}\PYG{n}{groupby}\PYG{p}{(}\PYG{l+s+s1}{\PYGZsq{}month\PYGZsq{}}\PYG{p}{)}\PYG{o}{.}\PYG{n}{sum}\PYG{p}{()}\PYG{o}{.}\PYG{n}{T}

\PYG{c+c1}{\PYGZsh{} The execution output is:}
\PYG{n}{\PYGZus{}\PYGZus{}output\PYGZus{}\PYGZus{}}\PYG{p}{:}
\PYG{n}{pandas}\PYG{o}{.}\PYG{n}{core}\PYG{o}{.}\PYG{n}{frame}\PYG{o}{.}\PYG{n}{DataFrame}

\PYG{c+c1}{\PYGZsh{} The user intent is:}
\PYG{n}{List} \PYG{n}{the} \PYG{n}{worldwide} \PYG{n}{gross} \PYG{n}{of} \PYG{n}{the} \PYG{n}{films} \PYG{n}{released} \PYG{k}{for} \PYG{n}{each} \PYG{n}{month} \PYG{n}{since} \PYG{n}{November}\PYG{p}{,}
\PYG{l+m+mi}{2021} \PYG{n}{to} \PYG{n}{June} \PYG{l+m+mf}{2022.} \PYG{n}{List}  \PYG{n}{the} \PYG{n}{months} \PYG{o+ow}{in} \PYG{n}{chronological} \PYG{n}{order}\PYG{o}{.}

\PYG{c+c1}{\PYGZsh{} The I/O specification is:}
\end{Verbatim}
\caption{The prompt used to query LLM for I/O summary.}
\end{subfigure}

\begin{subfigure}[t!]{1.0\textwidth}
\centering
\begin{Verbatim}[commandchars=\\\{\}]
\PYG{c+c1}{\PYGZsh{} In[ ]:}
\PYG{k+kn}{from} \PYG{n+nn}{datetime} \PYG{k+kn}{import} \PYG{n}{datetime}
\PYG{k+kn}{import} \PYG{n+nn}{pandas} \PYG{k}{as} \PYG{n+nn}{pd}
\PYG{n}{df} \PYG{o}{=} \PYG{n}{pd}\PYG{o}{.}\PYG{n}{read\PYGZus{}csv}\PYG{p}{(}\PYG{l+s+s1}{\PYGZsq{}dataset/bollywood\PYGZus{}2022.csv\PYGZsq{}}\PYG{p}{)}
\PYG{c+c1}{\PYGZsh{} In[ ]:}
\PYG{c+c1}{\PYGZsh{} Schema of Dataframes:}
\PYG{c+c1}{\PYGZsh{} Columns in df with example values:}
\PYG{c+c1}{\PYGZsh{} Movie (JugJugg Jeeyo), Worldwide (74.5), India Net (50.24), India Gross (54.5),}
\PYG{c+c1}{\PYGZsh{} Overseas (20.0), Movie Type (Bollywood), Released Date (24\PYGZhy{}Jun\PYGZhy{}22)}
\PYG{c+c1}{\PYGZsh{} In[ ]:}
\PYG{n}{df}\PYG{o}{.}\PYG{n}{columns} \PYG{o}{=} \PYG{p}{[}\PYG{n}{column}\PYG{o}{.}\PYG{n}{replace}\PYG{p}{(}\PYG{l+s+s1}{\PYGZsq{} \PYGZsq{}}\PYG{p}{,} \PYG{l+s+s1}{\PYGZsq{}\PYGZsq{}}\PYG{p}{)} \PYG{k}{for} \PYG{n}{column} \PYG{o+ow}{in} \PYG{n}{df}\PYG{o}{.}\PYG{n}{columns}\PYG{p}{]}

\PYG{c+c1}{\PYGZsh{} In[ ]:}
\PYG{n}{List} \PYG{n}{the} \PYG{n}{worldwide} \PYG{n}{gross} \PYG{n}{of} \PYG{n}{the} \PYG{n}{films} \PYG{n}{released} \PYG{k}{for} \PYG{n}{each} \PYG{n}{month} \PYG{c+c1}{\PYGZsh{} since}
\PYG{n}{November}\PYG{p}{,} \PYG{l+m+mi}{2021} \PYG{n}{to} \PYG{n}{June} \PYG{l+m+mf}{2022.} \PYG{n}{List} \PYG{n}{the} \PYG{n}{months} \PYG{o+ow}{in} \PYG{n}{chronological} \PYG{n}{order}\PYG{o}{.}
\PYG{l+s+sd}{\PYGZdq{}\PYGZdq{}\PYGZdq{}}
\PYG{l+s+sd}{Input\PYGZhy{}output Summary:}
\PYG{l+s+sd}{\PYGZus{}\PYGZus{}output\PYGZus{}\PYGZus{}: a pandas.core.DataFrame. Given the user intent and the code,}
\PYG{l+s+sd}{the salient columns (at most given 3) in the input dataframe are Worldwide,}
\PYG{l+s+sd}{Released Date. The output dataframe has columns (at most given 3) such as month,}
\PYG{l+s+sd}{Worldwide. Here is my code solution:}
\PYG{l+s+sd}{\PYGZdq{}\PYGZdq{}\PYGZdq{}}
\PYG{c+c1}{\PYGZsh{} In[ ]:}
\end{Verbatim}
\caption{A test example on \arcade/ augmented with a LLM generated I/O summary.}
\end{subfigure}

\begin{subfigure}[t!]{1.0\textwidth}
\centering
\begin{Verbatim}[commandchars=\\\{\}]
| April, 2022 (float) | December, 2021 (float) | February, 2022 (float)
| January, 2022 (float) | June, 2022 (float) | March, 2022 (float)
| May, 2022 (float) | November, 2021 (float) |
|\PYGZhy{}\PYGZhy{}\PYGZhy{}\PYGZhy{}\PYGZhy{}\PYGZhy{}\PYGZhy{}\PYGZhy{}\PYGZhy{}\PYGZhy{}\PYGZhy{}\PYGZhy{}\PYGZhy{}\PYGZhy{}\PYGZhy{}\PYGZhy{}\PYGZhy{}\PYGZhy{}\PYGZhy{}\PYGZhy{}\PYGZhy{}\PYGZhy{}\PYGZhy{}\PYGZhy{}\PYGZhy{}\PYGZhy{}\PYGZhy{}\PYGZhy{}\PYGZhy{}\PYGZhy{}\PYGZhy{}\PYGZhy{}\PYGZhy{}\PYGZhy{}\PYGZhy{}\PYGZhy{}\PYGZhy{}\PYGZhy{}\PYGZhy{}\PYGZhy{}\PYGZhy{}\PYGZhy{}\PYGZhy{}\PYGZhy{}\PYGZhy{}\PYGZhy{}\PYGZhy{}\PYGZhy{}\PYGZhy{}\PYGZhy{}\PYGZhy{}\PYGZhy{}\PYGZhy{}\PYGZhy{}\PYGZhy{}\PYGZhy{}\PYGZhy{}\PYGZhy{}\PYGZhy{}\PYGZhy{}\PYGZhy{}|
| 1748.13 | 10988.6 | 1812.99 | 114.12 | 4763.05 | 3849.25 | 7730.75 | 169.23 |
\end{Verbatim}
\caption{The ground truth dataframe of the output in the above test example.}
\end{subfigure}
\captionof{listing}{An noisy example of LLM generated I/O summary. The LLM generated I/O specification is inaccurate as evident from the discrepancies between the I/O summary in part \textbf{(b)} and the ground truth in part \textbf{(c)}.}
\label{fig:appendix:IOsummary_ex2}
\end{figure}

\begin{table}[t]
    \small
    \centering
    \begin{tabular}{@{}lcccc}
    \toprule
     \multirow{3}{*}{\textbf{Methods}} & 
     \multicolumn{4}{c}{\textbf{\arcade/}} \\
     & \multicolumn{2}{c}{\passat{5}} & \multicolumn{2}{c}{\passat{20}} \\
      \cmidrule(lr){2-3} \cmidrule(lr){4-5}
      & \textrm{No Context} & \textrm{Full Context} & \textrm{No Context} & \textrm{Full Context} \\ 
     \rowcolor{black!10!} \multicolumn{5}{c}{\textbf{\textit{\giftforcode/}}} \\
    Code LLM (no spec.) & \textbf{20.78} & \textbf{34.33} & \textbf{33.40} & \textbf{46.94} \\
    ~~$+$ \typedesc/ & 20.59 & 33.43 & 32.98 & 46.01 \\
    ~~$+$ \ioexample/ & 20.04 & 31.76 & 31.40 & 43.60 \\
    ~~$+$ \iosummary/ & 20.04 & 32.39 & 32.24 & 45.14 \\
    \bottomrule
    \end{tabular}
    \caption{\passat{k} on \arcade/. For each type of I/O specification in \cref{tab:appendix:exp:arcade_result} (\eg~$+$\iosummary/), intents are augmented with I/O specifications of that type (\eg~intents inline with I/O summary) in fine-tuning data. At test time, input intents do not contain have any specification.}
    \label{tab:appendix:exp:arcade_result}

\end{table}

\end{document}